\documentclass[letterpaper, 10 pt, conference]{ieeeconf}  

\IEEEoverridecommandlockouts                              

\usepackage{times} 
\usepackage{amsmath} 
\usepackage{amssymb}  
\usepackage{color}
\usepackage{booktabs}
\usepackage{graphicx}
\usepackage{subfigure}
\usepackage{multirow,booktabs}
\usepackage{stfloats} 
\usepackage{fancyhdr}
\usepackage{hyperref}
\usepackage{footnote}
\usepackage{float}

\newcommand{\bp}{\mathbf{p}}

\newcommand{\bo}{\mathbf{o}}

\newcommand{\bT}{\mathbf{T}}

\newcommand{\bd}{\mathbf{d}}

\newcommand{\cR}{\mathcal{R}}

\makeatletter
\DeclareRobustCommand\onedot{\futurelet\@let@token\@onedot}
\def\@onedot{\ifx\@let@token.\else.\null\fi\xspace}

\makeatother

\renewcommand{\eqref}[1]{Eq.~\ref{#1}}

\newif\ifcomment
\commenttrue
\ifcomment
	\newcommand{\ag}[1]{ \noindent {\color{red} {\bf Andreas:} {#1}} }
	\newcommand{\yl}[1]{ \noindent {\color{cyan} {\bf Yiyi:} {#1}} }

\else
	\newcommand{\ag}[1]{}
	\newcommand{\yl}[1]{}
\fi

\title{\LARGE \bf
NF-Atlas: Multi-Volume Neural Feature Fields\\ for Large Scale LiDAR Mapping
}

\author{Xuan Yu$^1$, Yili Liu$^1$, Sitong Mao$^2$, Shunbo Zhou$^2$, Rong Xiong$^1$, Yiyi Liao$^1$, Yue Wang$^1$  
\thanks{Xuan Yu, Yili Liu, Rong Xiong and Yue Wang are with the State Key Laboratory of Industrial Control Technology and Institute of Cyber-Systems and Control, Zhejiang University, Hangzhou, China. Yiyi Liao is with College of Information Science and Electronic Engineering, Zhejiang University, Hangzhou, China. Sitong Mao and Shunbo Zhou are with Huawei Cloud Computing Technologies Co., Ltd., Shenzhen, China. }
}

\begin{document}

\maketitle
\thispagestyle{empty}
\pagestyle{empty}

\begin{abstract}
LiDAR Mapping has been a long-standing problem in robotics. Recent progress in neural implicit representation has brought new opportunities to robotic mapping. In this paper, we propose the multi-volume neural feature fields, called NF-Atlas, which bridge the neural feature volumes with pose graph optimization. By regarding the neural feature volume as pose graph nodes and the relative pose between volumes as pose graph edges, the entire neural feature field becomes both \textit{locally rigid} and \textit{globally elastic}. Locally, the neural feature volume employs a sparse feature Octree and a small MLP to encode the signed distance function (SDF) of the submap with an option of semantics. Learning the map using this structure allows for end-to-end solving of maximum a posteriori (MAP) based probabilistic mapping. Globally, the map is built volume by volume independently, avoiding catastrophic forgetting when mapping incrementally. Furthermore, when a loop closure occurs, with the elastic pose graph based representation, only updating the origin of neural volumes is required without remapping. Finally, these functionalities of NF-Atlas are validated. Thanks to the sparsity and the optimization based formulation, NF-Atlas shows competitive performance in terms of accuracy, efficiency and memory usage on both simulation and real-world datasets. The project page is: \href{https://yuxuan1206.github.io/NFAtlas/}{https://yuxuan1206.github.io/NFAtlas/}
\end{abstract}


\section{Introduction}


Mapping is a fundamental task for robotics. In particular, dense mapping is essential to many robotic applications e.g. navigation. For large-scale outdoor scenes, it is challenging to achieve fast and accurate dense mapping while maintaining a low memory cost.

LiDAR has emerged as a popular sensor for mapping due to its ability to provide range measurements. Conventional LiDAR-based methods can achieve fast mapping with a relatively low memory cost, adopting a sparse, discretized volume representation, e.g., Octree~\cite{octomap} or hash table~\cite{hash}. However, the reconstructed map of conventional methods may be noisy and contain many holes. This is largely due to the fact that the voxels are assumed to be independent of each other~\cite{kinectfusion,octomap} to enable efficient sequential fusion. However, the independence assumption yields a crude approximation of the real scene and leads to non-smooth results, thus hampering the fusion accuracy. 


\begin{figure}[!htbp]
\centering
\includegraphics[width=\linewidth,keepaspectratio]{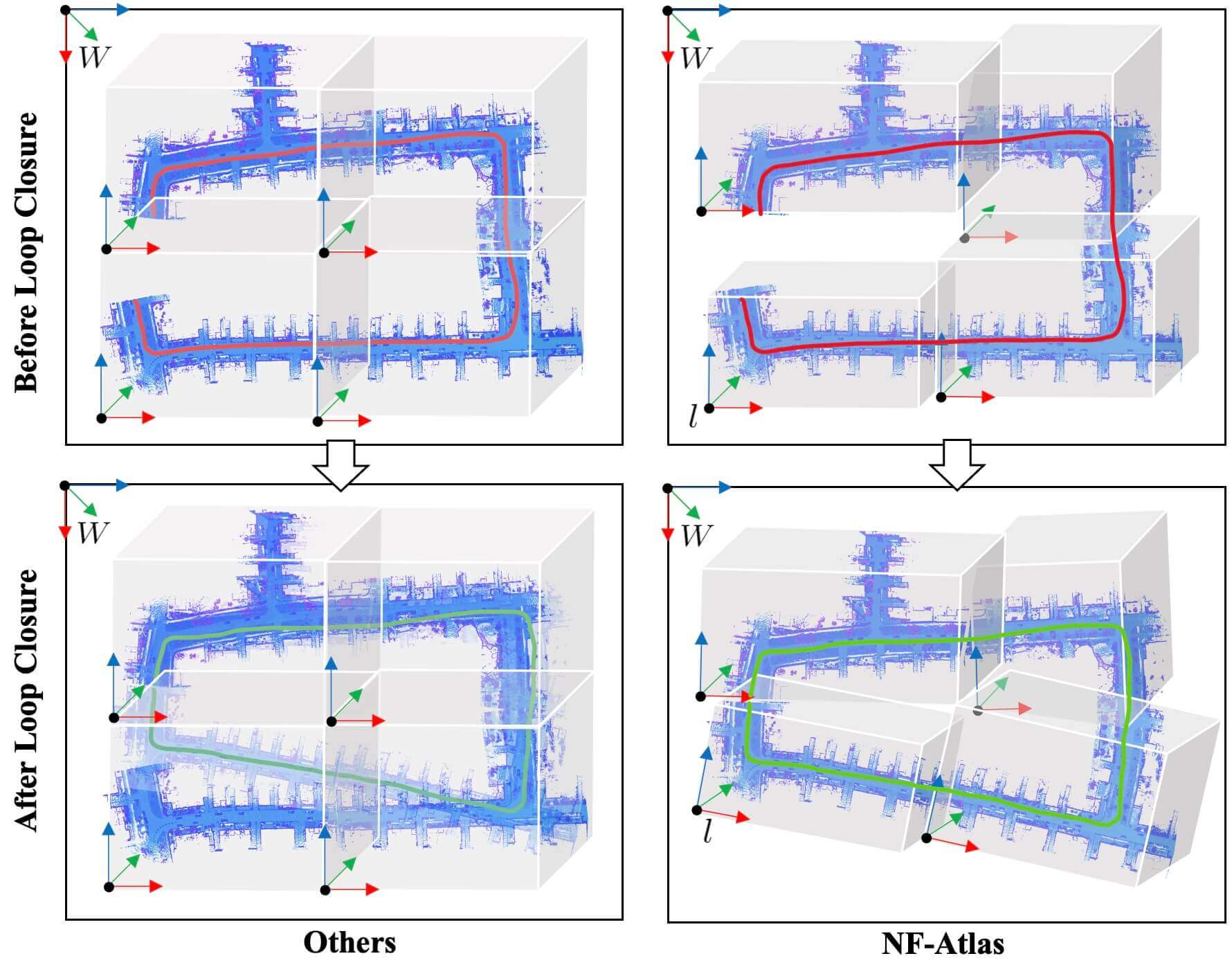}\\
\vspace{-0.3cm}
\caption{\label{idea}The map is divided into multiple submaps. Each submap is a neural features volume, which can be fixed to world coordinates $W$ (left column) and  anchor pose coordinates $l$ (right column). When a loop closure happens, the map in the world coordinates fixed submap calls for remapping due to the trajectory correction, while the anchor pose coordinates fixed submap calls for only a volume transform determined by the trajectory correction.}
\vspace{-0.5cm}
\end{figure}

Recent years witnessed a great success of implicit neural representations~\cite{park2019deepsdf,mildenhall2021nerf,mescheder2019occupancynetworks} that encode scenes using coordinate-based multi-layer perception (MLP)~\cite{mildenhall2021nerf,zhu2022nice}. In particular, neural radiance fields (NeRF) propose to combine coordinate-based MLPs with differentiable volume rendering in an end-to-end manner. In this way, NeRF actually formulates mapping as a maximum likelihood optimization given the scene observations. This formulation allows for further adding regularizations, achieving the high-quality view synthesis and the low memory utilization~\cite{yu2021plenoxels,sun2022neuralinthewild,azinovic2022neuralrgbd,wang2022gosurf,ortiz2022isdf}. Some works extend NeRF to integrate range measurements, demonstrating an impressive quality in small-scale environments e.g. rooms~\cite{azinovic2022neuralrgbd, wang2022gosurf, ortiz2022isdf}. 
However, when applying to large-scale outdoor mapping, the large MLP and slow rendering make the training difficult~\cite{rematas2022urban, shi2022city}. Efforts have been made to improve the training efficiency by representing the map as a 3D feature volume in combined with a small MLP, but such dense representation consumes very high memory when the environment is large~\cite{rematas2022urban, shi2022city}.


Further challenges arise when the map is required to be built incrementally in some applications. First, the implicit neural representations suffer from catastrophe forgetting. Moreover, if the trajectory is also updated incrementally, both conventional mapping and neural mapping methods call for total re-integration or re-optimization~\cite{pan2020gem,wang2021elastic}. These difficulties raise a question: \textit{what is the representation suitable for large scale LiDAR mapping}? 

\begin{figure*}[htbp]
\centering
\includegraphics[width=\textwidth,keepaspectratio]{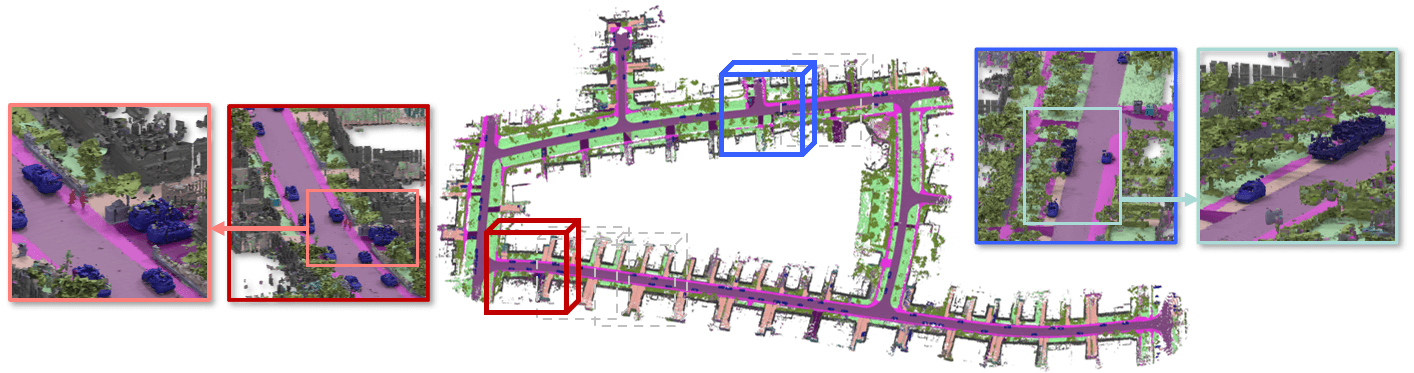}\\
\vspace{-0.3cm}
\caption{\label{hole} The center image shows the semantic reconstruction of the global map using LiDAR and semantic labels on Seq.00 of the KITTI-360. Several example volumes are shown in the map, among which the red and blue ones are highlighted with the surface details in the side images.}
\vspace{-7mm}
\end{figure*}

In this paper, we propose to represent a large-scale map as multi-volume neural feature fields, named NF-Atlas, bridging the advantages of sparse volume representation and implicit neural representations. Our key idea is to represent a large map using multiple submaps connected by a pose graph, where each submap is a sparse, multi-scale neural feature volume encoding the signed distance function (SDF). Therefore, as shown in Fig.~\ref{idea}, the neural feature fields as a whole is \textit{locally rigid} and \textit{globally elastic}. In local, we obtain an Octree using the LiDAR observations and then model the neural feature volume as a multi-scale feature Octree, leading to a lightweight representation. Further, the multi-scale Octree representation allows for modeling details with better accuracy and efficient sampling. On the other hand, we formulate the mapping as maximum a posteriori problem, which can be optimized in an end-to-end manner. The semantic cues can also be easily incorporated in the optimization. In global, our formulation does not suffer from catastrophe forgetting when mapping incrementally, as each submap is modeled independently. Even in the case of loop closure, only the origins of the submaps need to be updated base on the pose graph optimization owing to the elastic inter-submap connection, 
while the local area's poses and mapping can be considered invariant. In the experiments on both simulation and real world datasets, we show that NF-Atlas achieve better accuracy and efficiency with a relatively low memory consumption than the comparative learning-based and conventional mapping methods. A semantic reconstruction of the urban environment is shown in Fig.~\ref{hole}. In summary, the contributions involve:
\begin{itemize}
    \item A sparse neural feature volume formulating LiDAR mapping as a maximum a posteriori problem, which is end-to-end optimized for better quality.
    \item An atlas organizing multiple neural feature volumes by a pose graph, which forms an incremental elastic feature fields that avoids catastrophe forgetting and remapping.
    \item Experiments on both simulation and real-world datasets validate the advantages of NF-Atlas. The CUDA implementation of regularizer back-propagation are released.
\end{itemize}

\section{Related Works}
\subsection{Learning-free Reconstruction}
Map representations calls community's attention for long years. The SDF stores the closest distance from each point to the surface of the object, which is often used as an implicit representation of the object to represent the surface details. Newcombe et al.~\cite{kinectfusion} popularized with KinectFusion use the truncated signed distance function (TSDF) for mapping. However, KinectFusion is limited to RGB-D-based indoor reconstruction due to the dense volume. Whelan et al.~\cite{whelan2015Real-time} propose a fused volumetric method for globally consistent surface reconstructions to achieve spatially extended mapping. Subsequently, sparse voxels~\cite{voxblox,voxfield,vdbfusion} are applied to the scalable mapping system. Besides, some approaches are also achieved by combining sparse volume octree with occupancy grid map~\cite{vespa2018Efficient_octree,wang2021elastic}. Suma++~\cite{suma++} provide surfel-based mapping and accurate odometry, without following a pre-defined grid. In addition, several works~\cite{vineet2015incrementaldensesemanticstereo,schmid2022panopticmulti-tsdfs,suma++} integrate semantic information to facilitate the mapping process. When a loop closure occurs, such dense representation calls for remapping along the corrected trajectory. To save the computation,  ~\cite{pan2020gem} propose to organize multiple dense maps as a pose graph, which simplifies the remapping to coordinate transform.

\subsection{Learning-based Reconstruction}
Neural fields become a competitive representation for reconstruction recently ~\cite{mildenhall2021nerf,mescheder2019occupancynetworks,park2019deepsdf}. Sdfdiff~\cite{jiang2020sdfdiff} and DIST~\cite{liu2020dist} address differentiable sphere tracing to learn SDFs of 3D objects from images. DVR~\cite{DVR} and IDR~\cite{yariv2020multiview} determine the radiance directly on the surface of an object and provide a differentiable rendering formulation, but requiring foreground mask as supervision. Wang et al.~\cite{wang2021neus} develop a new volume rendering method to train a bias-free neural SDF representation, which, however, takes a long time, failing to reconstruct the featureless surfaces. Along this direction, Sun et al.~\cite{sun2022neuralinthewild} focus on more efficient sampling strategy that enables accurate surface reconstruction. In order to accelerate the training speed, some methods employ hierarchical representations, say octree ~\cite{takikawa202NGLOD} and multi-resolution grids ~\cite{muller2022instantngp,sun2022directvoxelgrid,yu2021plenoxels}. However, these methods are not evaluated on large scale environments due to the memory and efficiency.

\subsection{Learning-based Incremental Reconstruction}
Current research efforts consider incremental mapping of implicit representations as the continual learning about the neural fields. With respect to the implicit reconstruction of indoor scenes using RGB-D sensors, iMAP~\cite{sucar2021imap} presents the first implicit SLAM based on neural radiation fields. NICE-SLAM~\cite{zhu2022nice} improves upon iMAP by incorporating a pre-trained geometric prior. iSDF~\cite{ortiz2022isdf} employs a neural network to regress the input 3D coordinates to the signed distance. Azinović et al.~\cite{azinovic2022neuralrgbd} augments the NeRF framework with TSDF to represent the surface instead of the volume. In the context of incremental implicit reconstruction using urban sparse LiDAR point clouds, Shi et al.~\cite{shi2022city} proposes an SDF-based semantic mapping approach that utilizes a three-layer sampling strategy and panoptic representation to mitigate catastrophic forgetting during incremental reconstruction. These works all replay keyframes from historical buffer and train the network with current observations together. However, the past mapping still degenerates when the learning time is bounded. In addition, these methods do not consider the loop closure explicitly, which may bring significant trajectory correction, causing the remapping.

\section{Optimization based Mapping}

The NF-Atlas contains several submaps, each of which models the environment rigidly as an implicit function that maps a coordinate to a signed distance value using a neural feature volume. We state the neural feature volume based mapping as an optimization problem, which jointly optimizes volumes parameters and optionally the poses. 

\subsection{Neural Feature Volume}

\begin{figure}[t]
\flushleft
\includegraphics[width=8.8cm,keepaspectratio]{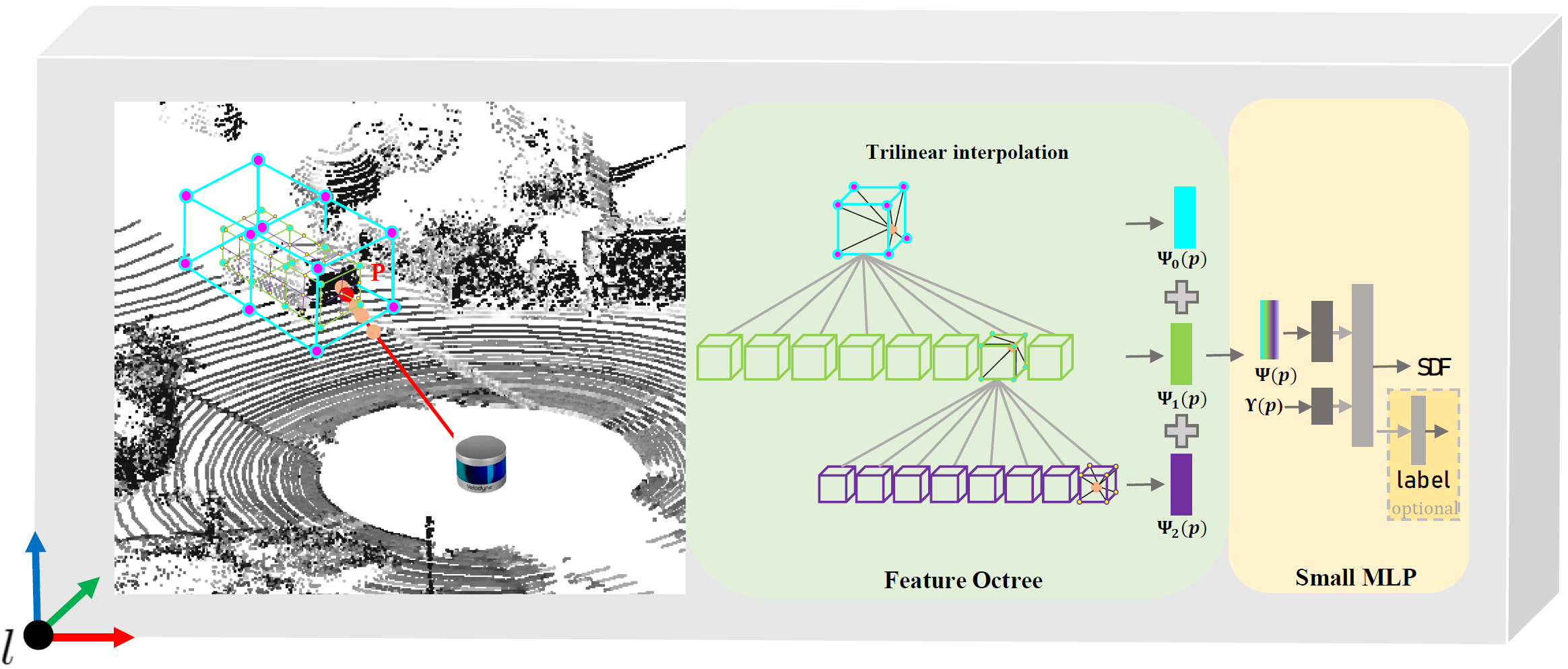} 
\vspace{-6mm}
\caption{\label{NF}The architecture of the neural feature volume with an optional semantic branch which is described in detail in Sec.III.}
\vspace{-6mm}
\end{figure}

The architecture of neural feature volume is shown in Fig.~\ref{NF}, which consists of a feature Octree and a small MLP, yielding the SDF of a query 3D point.

\textbf{Feature Octree:} The feature Octree maps a query 3D point $\bp\in \mathcal{R}^3$ to a feature vector $\Psi(\bp) \in \mathcal{R}^N$. First, we employ an Octree based structure $\Psi$ to sparsely encode the local area into features. Specifically, as shown in~\cite{takikawa202NGLOD}, each Octree node stores a learnable feature $\psi_{i,j}\in \mathcal{R}^N$, where $i$ indicates the tree level, while $j$ indicates the index. Given a query point $\bp$, its feature is acquired by summing the tri-linearly interpolated features in top $K$ levels as:
\begin{equation}
    \Psi(\bp)= \sum_{i=0}^{K-1} triInterp(\psi_{i,j\in \mathcal{N}(\bp)})
\end{equation}
where $triInterp$ denotes the tri-linear interpolation, $\mathcal{N}(\bp)$ is the set of the eight nearest Octree nodes i.e. corners of a cube containing $\bp$. The Octree-based sparse volume achieves a balance between reconstruction quality and efficiency through representing 3D shapes in a compressed format which stores multi-level features. 



\textbf{Small MLP:} We adopt a small MLP to map the feature vector $\Psi(\bp)$ and the input coordinate $\bp$ to an SDF value: 
\begin{equation}
    s = f_\theta(\Psi(\bp),\gamma(\bp))
    \label{sdf}
\end{equation}
where $s\in \cR$ is the SDF value of $\bp$, $\gamma(\cdot)$ denotes position encoding that is applied to each element of $\bp$:
\begin{equation}
\small{
\setlength{\arraycolsep}{1.2pt}
    \gamma(\bp) = \begin{bmatrix}
           \cos(2^0\pi \bp), & \sin(2^0\pi \bp), &\dots & \cos(2^{L-1}\pi \bp), & \sin(2^{L-1}\pi \bp)
         \end{bmatrix}
 }
\end{equation}
Different from the MLP in original NeRF~\cite{mildenhall2021nerf}, $f_\theta$ is much smaller owing to the octree feature, thus highly efficient while still keeping the quality of the reconstruction.   

\subsection{Differentiable Range Approximation}
Given SDF of a query point, we can find the surface in the field easily. However, at the early stage of the training, the SDF is far from convergence. Therefore, we introduce the differentiable range approximation from the neural feature volume following the differentiable volume rendering in NeuS~\cite{wang2021neus}. Given the origin and direction of a ray, denoted by $(\bo,\bd)$, we can represent a spatial point along the ray as $\bp = \bo + \rho \bd$. By sampling $\{\rho_n\}$, we have a batch of points on the ray $\{\bp_n\}$ arranged in near-to-far order. We can query their corresponding SDFs $\{s_n\}$ to approximate the range $r$ along the ray as:
\begin{equation}
    r = \sum_{n=1}^{N}{T_{n} \alpha_{n} \rho_{n}}
\end{equation}
where $T_{n}=\prod_{m=1}^{n-1}(1-\alpha_{m})$, $\alpha_{n}$ is the discrete opacity value defined under the S-density function assumption~\cite{wang2021neus} as:
\begin{equation}
    \alpha_{n} = \max\left(\frac{ \Phi(s_n)-\Phi(s_{n+1}) }{ \Phi(s_n) }, 0\right) 
\end{equation}
where $\Phi(x)$ is Sigmoid function $\Phi(x)=(1+e^{-\xi x})^{-1}$ with a temperature coefficient $\xi$.

\textbf{Sampling:} Thanks to the sparse Octree and the point cloud, we can accurately sample the points by voxel-guided sampling and surface-guided sampling from the near-surface region~\cite{sun2022neuralinthewild}. By skipping the empty space along the ray, we can omit sampling in the low informative region, improving the sampling efficiency significantly.

\subsection{Likelihood Measurement Model}

\textbf{Range Measurements:} By stitching the neural feature volume and the differentiable range approximation, we follow the classical probabilistic mapping theory~\cite{thrun2002probabilistic} to design a range measurement model $p(r|\bo,\bd,\theta,\Psi)$
\begin{equation}
    p_{r}(r|\bo,\bd,\theta,\Psi) = \frac{1}{\sigma_r \sqrt{2\pi}} \exp\left\{-\frac{(\hat{r}-r(\bo,\bd))^2}{2\sigma_r^2}\right\}
    \label{rl}
\end{equation}
where $\sigma_r^2$ is the variance of the measurement noise, $r(\bo,\bd)$ is the approximated range along ray $(\bo,\bd)$ from the neural feature volume, $\hat{r}$ is the measured range along the ray. In this way, the measurement model reflects the likelihood of the measured range in the real world. The vital advantage of this model is the differentiable pathway to map parameters i.e. neural feature volume, even the pose i.e. ray parameters.



\textbf{SDF Measurements:} By formulating as a maximum likelihood problem, we can further add more measurements. In NF-Atlas, We also employ SDF measurement model following~\cite{ortiz2022isdf,azinovic2022neuralrgbd,wang2022gosurf}. Given a LiDAR point measurement with a range of $\hat{r}$, we consider a sample $\bp = \bo+\rho \bd$  as near-surface if $|\hat{r}-\rho|\leq \tau$. We derive the measured SDF of this point as $b = \hat{r}-\rho$. Then we define an truncated Laplace distribution for SDF measurement model:
\begin{equation}
    p_{near}(s|\bp,\theta,\Psi) = \frac{1}{2\lambda} \exp\left\{-\frac{| b -s(\bp)|}{\lambda}\right\}
\end{equation}
where $\lambda$ is the bandwidth coefficient, $s(\bp)$ is the approximated SDF of sample $\bp$ as (\ref{sdf}). For the far-surface sample $\bp$ satisfying $|\hat{r}-\rho|> \tau$, we define an exponential density~\cite{wang2021neus} as the measurement model:
\begin{equation}
    p_{far}(s|\bp,\theta,\Psi) =  \eta\exp\left\{-\max(0, e^{-\beta s(\bp)}-1, s(\bp) - b ) \right\}
\end{equation}
where $\eta$ is a normalizer for the validness of the density. As a whole, the likelihood of the measured SDF is:
\begin{equation}
    p_s(s|\bp,\theta,\Psi)=\begin{cases}
                    p_{near}(s|\bp,\theta,\Psi) & |b|\leq \tau\\
                    p_{far}(s|\bp,\theta,\Psi) & o.w.
                    \end{cases}
                    \label{sl}
\end{equation} 

\textbf{Semantic Measurements:} We further add semantic measurements into our mapping system as~\cite{semanticnerf}. We augment (\ref{sdf}) with semantic prediction as:
\begin{equation}
    s,p(c) = f_\theta(\Psi(\bp),\gamma(\bp))
    \label{semantic}
\end{equation}
where $p(c)$ is a multinomial distribution of class $c$ implemented by a softmax branch in parallel with the SDF prediction. We approximate the semantics of the ray as:
\begin{equation}
    l(c) = \sum_{n=1}^{N}{T_{n} \alpha_{n} p(c_{n})}
\end{equation}
Then the likelihood of the measured semantic class $\hat{c}$ is:
\begin{equation}
    p_c(c = \hat{c}|\bo,\bd,\theta,\Psi) = softmax(l(c)) |_{c=\hat{c}}
    \label{cl}
\end{equation} 
which is the predicted probability of the measured class. 

\subsection{Probabilistic Mapping}

With measurement models above, we arrive at the data likelihood as a product of (\ref{rl}), (\ref{sl}) and (\ref{cl}), upon which we further add prior as regularization to derive the posterior. As a result, we formulate probabilistic LiDAR mapping as a maximum a posteriori (MAP) problem:
\begin{equation}
    \tilde{\theta},\tilde{\Psi},\tilde{\bo},\tilde{\bd} =\arg\max  \prod \underbrace{p_r \cdot p_s \cdot p_c}_{\text{Likelihood}} \cdot \underbrace{p_e \cdot p_h}_{\text{Prior}} 
    \label{total}
\end{equation}
where $p_e$ is a prior on the identity gradient of the SDF, $p_h$ is a prior on the smoothness of the neighborhood in the real world environment, the product $\prod$ means multiplication across all samples along all rays. The MAP formulation reserves the differentiable pathway from all terms to map parameters i.e. neural feature volume, and leaves the pose i.e. ray parameters from only likelihood terms, as we apply no prior on the mapping trajectory.

\textbf{Prior Terms:} Specifically, we have $p_e$ as
\begin{equation}
    p_{e}(\theta,\Psi) = \frac{1}{\sigma_e \sqrt{2\pi}} \exp\left\{-\frac{(1-\|\nabla_{\bp} s(\bp)\|)^2}{2\sigma_e^2}\right\}
    \label{pe}
\end{equation}
where $\sigma_e^2$ is the variance. For $p_h$, we have the smoothness prior as~\cite{yu2021plenoxels,wang2022gosurf}:
\begin{equation}
    p_{h}(\theta,\Psi) = \frac{1}{\sigma_h \sqrt{2\pi}} \exp\left\{-\frac{\| \nabla_{\bp} s(\bp) - \nabla_{\bp} s(\bp+\Delta \bp) \|^2}{2\sigma_h^2}\right\}
    \label{ph}
\end{equation}
where $\sigma_h^2$ is the variance, $\Delta \bp$ is a small perturbation vector of $\bp$ to evaluate the normal direction consistency.

\textbf{Implementation of Back-propagation:} The details of the derivation are shown in Appendix. 
Note that the back-propagation of the two prior regularizers requires the evaluation of 2nd-order derivatives. For fast computation of the double back-propagation of tri-linear interpolation in Octree, we implement the operation by CUDA based on~\cite{wang2022gosurf}.

\section{Large Scale Mapping as Atlas}

For the large-scale scene, we employ a pose graph to organize the multiple neural feature volumes as nodes that capture the local area. Each edge between the nodes indicates the measured relative pose between the two connected neural feature volumes. In this way, the origins of the volumes can be updated, making the whole map an elastic neural feature field. Like an atlas, we can check the map volume by volume, or combine them as a whole.

\subsection{Incremental Mapping} 
When mapping the environment, we build a neural feature volume using a sequence of poses and their measurements following (\ref{total}). The origin of the $l$th neural feature volume is fixed to the starting pose $\bT_{l}$ of the $l$th sequence.


We pick frames in each volume based on the move distance threshold and set the overlap between two neighboring volumes to keep a smooth transition. As the mapping progresses, we freeze the past volumes and incrementally initialize a new volume using the poses spanning a similar distance, as well as their measurements. In this way, the regions covered by the neural feature volumes can be similar, reserving the computation complexity bounded. In addition, as each volume is optimized separately, we avoid the catastrophic forgetting of the neural network.

\textbf{Loop Closure without Remapping:} As each neural volume is fixed to the starting pose, when a loop closure occurs, updating the origins of these volumes to match the revised robot trajectory $\bT_{l}$ is sufficient, and there is no need to adjust their volume parameters. This is the main advantage compared with existing large-scale neural mapping methods that fix the volumes to global area partitions, which calls for remapping when the trajectory is updated. 

\textbf{On-demand Global Mapping:} If a global consistent map is required, we can extract the map SDF from multiple neural feature volumes according to their most recent origins solved by pose graph optimization. Since the map is frozen after it is built, we only need to query the map. As shown in Fig.~\ref{idea}, given a point $\bp^W$ in world coordinate $W$, assume its closest volume is the $l$th volume, then its SDF is:
\begin{equation}
    s = f_{\theta_l}(\Psi_l(\bT_l^{-1} \bp^W),\gamma(\bT_l^{-1} \bp^W)) 
\end{equation}
where $\theta_l$ and $\Psi_l$ are the parameter of the $l$th volume. 

\subsection{Loop Closure Updating}
The loop detection is divided into coarse search and fine registration. We pick the nearest one of the volumes whose timestep is more than a certain step earlier than the current volume for fine registration. Then, we calculate the relative pose between the detected loop closure pair by aligning the two neural feature volumes. Say the two volumes are indexed by $l$ and $j$, they have an overlap since there exists a loop closure. Then we estimate the relative pose by maximizing the likelihood of measurements in volume $j$ conditioned on the volume $l$ as:
\begin{equation}
    \tilde{\bo}_j, \tilde{\bd}_j = \arg \max p_r(r_j|\bo_j,\bd_j,\theta_l,\Psi_l)
    \label{volumealign}
\end{equation}
Note that we do not update the map at the loop closure stage. Since $\tilde{\bo}_j$ and $\tilde{\bd}_j$ are parameterized by pose $\bT_{j}$, we arrive at the optimal relative pose, which is added to the pose graph as a new edge, activating the pose graph optimization. Compared with conventional LiDAR based 3D-3D point cloud registration, such inter-volume measurement based alignment utilizes the relationship between range measurements and the 3D SDF to guide the pose estimation.


\begin{figure*}[ht]
    \centering
    \subfigure{
        \includegraphics[width=3.4cm]{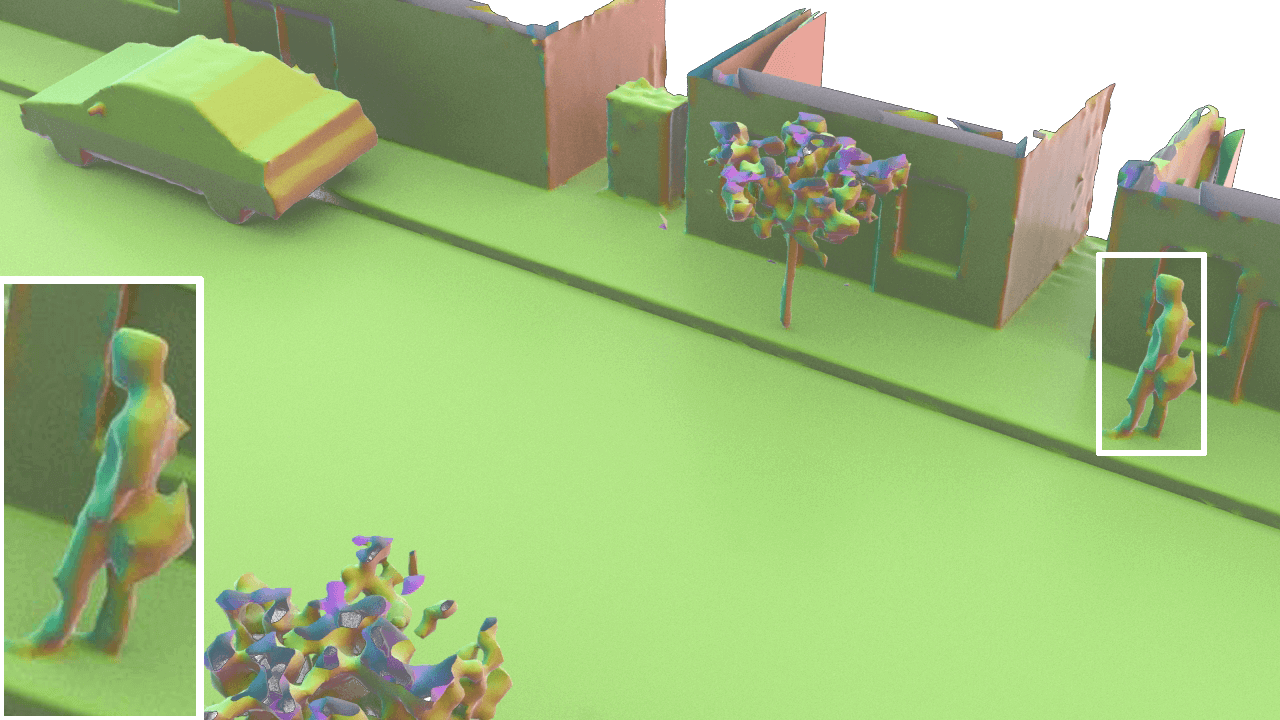} 
    }\hspace{-3.5mm}
    \subfigure{
        \includegraphics[width=3.4cm]{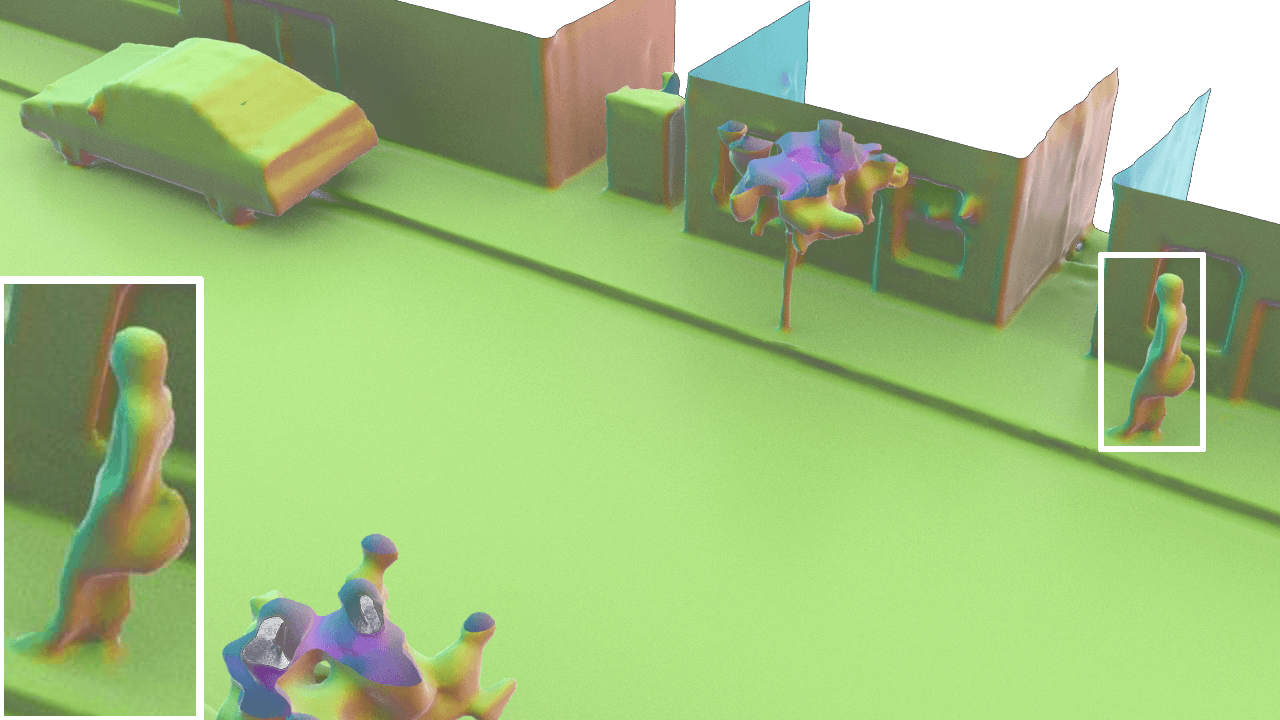} 
    }\hspace{-3.5mm}
    \subfigure{
        \includegraphics[width=3.4cm]{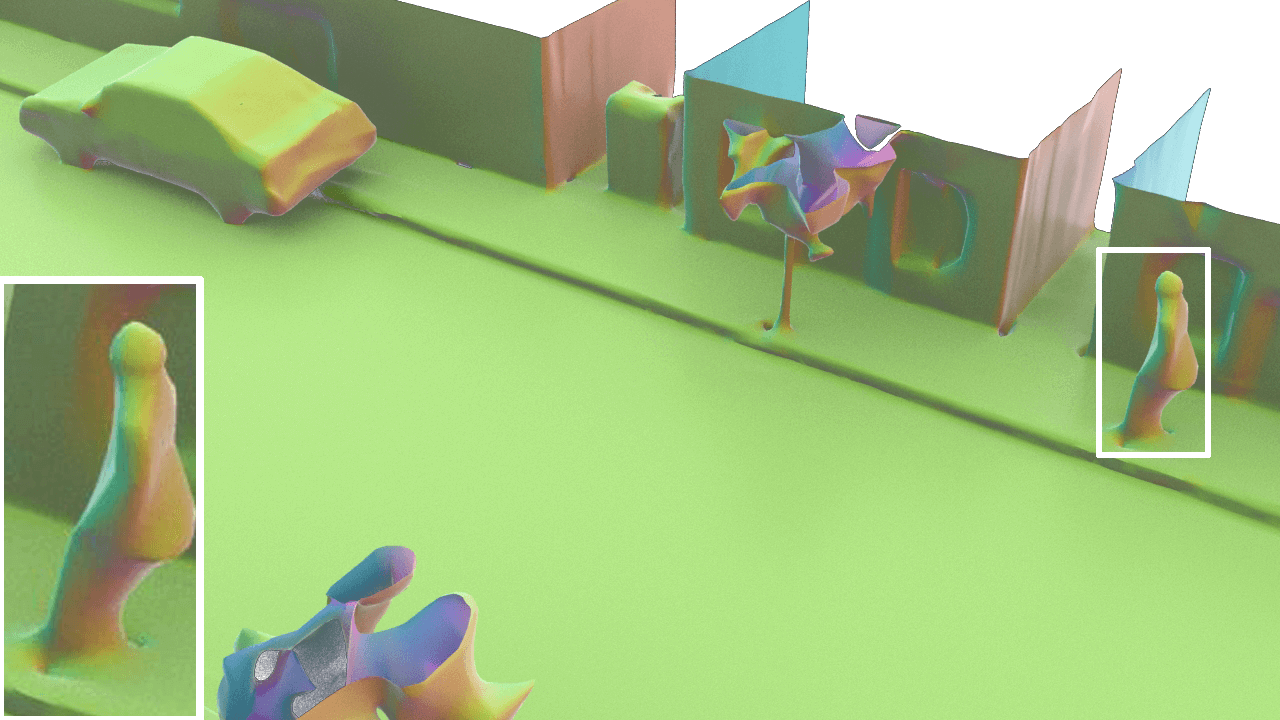}
    }\hspace{-3.5mm}
    \subfigure{
        \includegraphics[width=3.4cm]{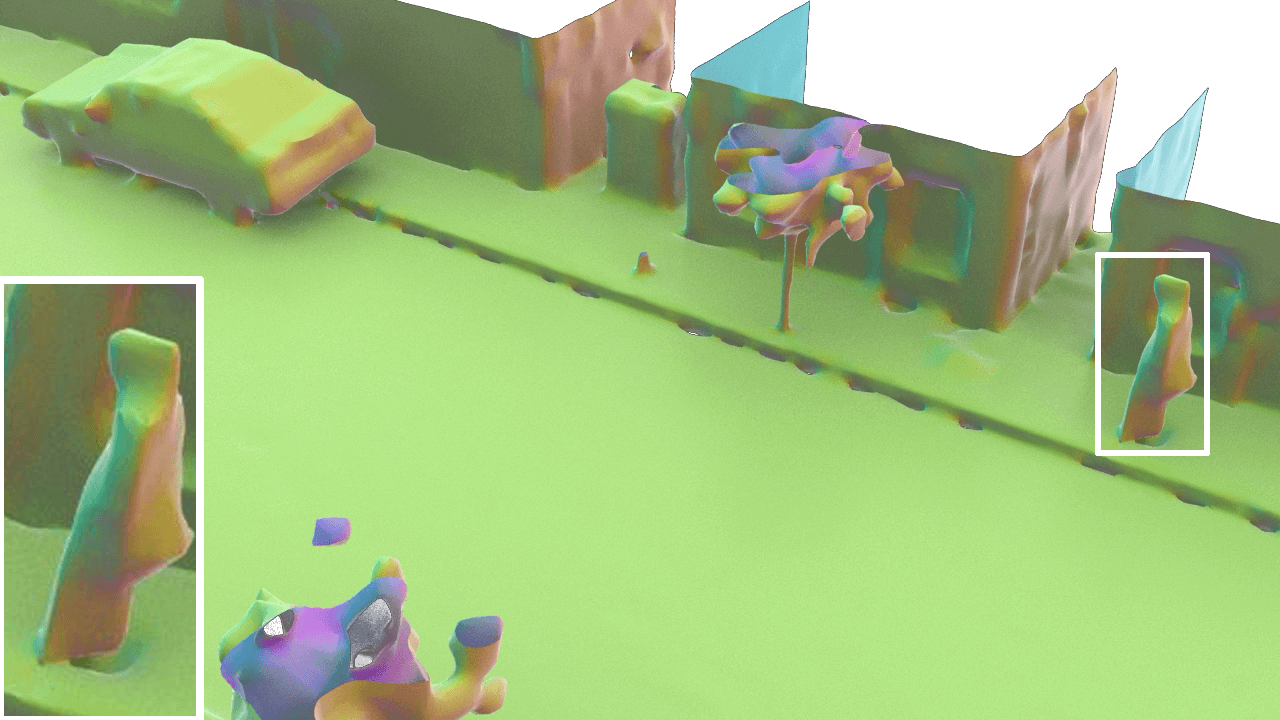}
    }\hspace{-3.5mm}
    \subfigure{
        \includegraphics[width=3.4cm]{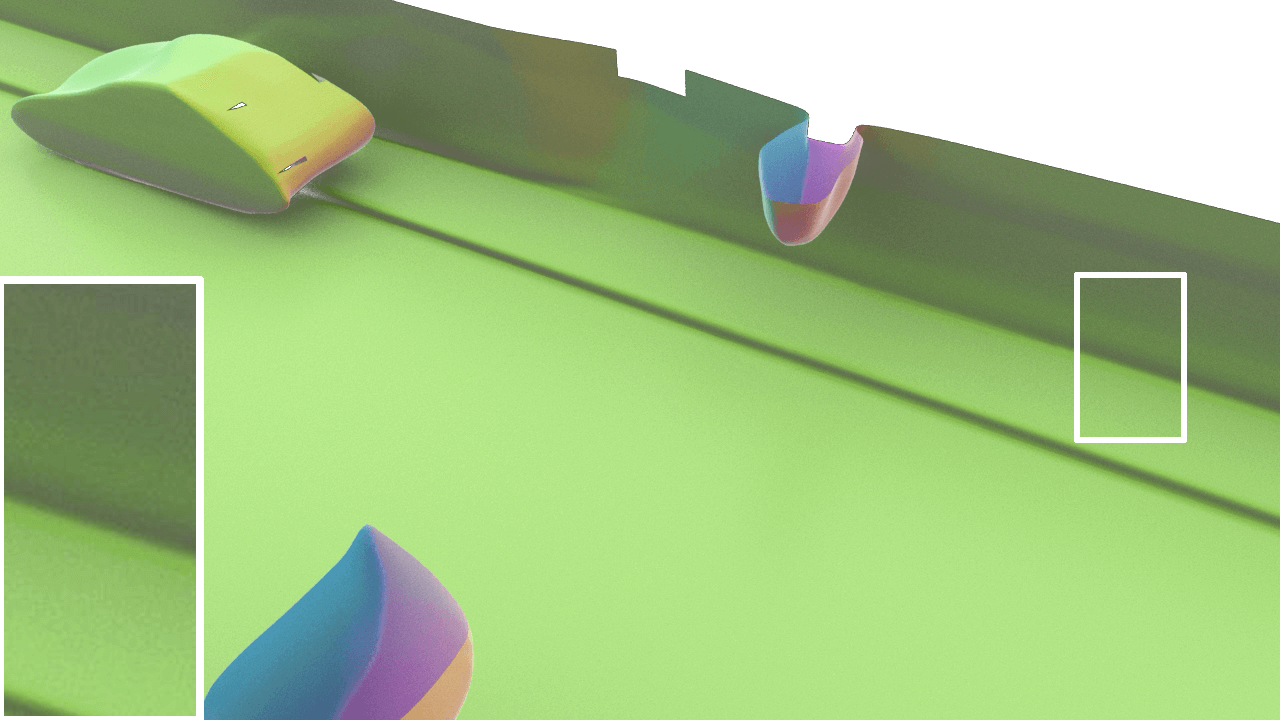}
    }

    \setcounter{subfigure}{0}
    \subfigure[Ours]{
        \includegraphics[width=3.4cm]{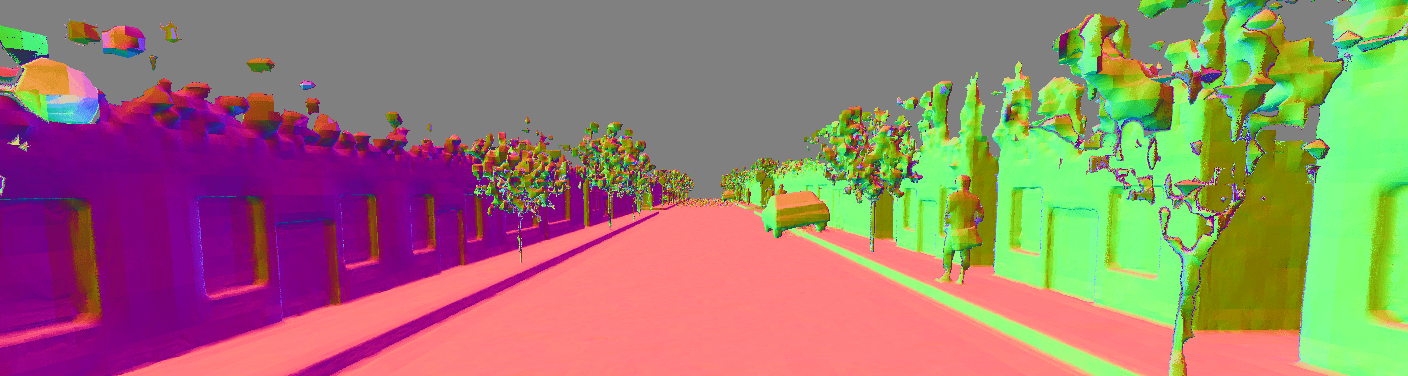}
    }\hspace{-3.5mm}
    \subfigure[Hash Grid + MLP]{
        \includegraphics[width=3.4cm]{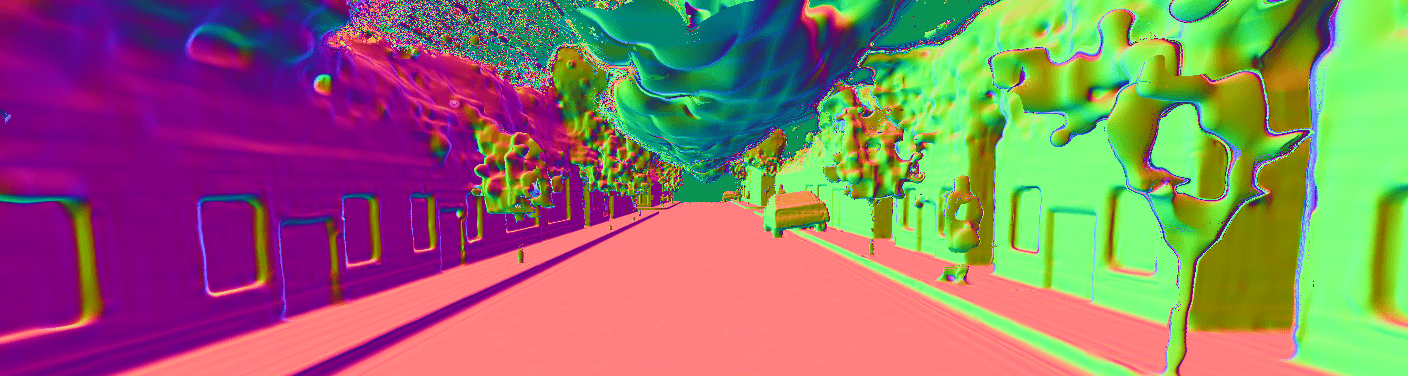} 
    }\hspace{-3.5mm}
    \subfigure[Dense Grid + MLP]{
        \includegraphics[width=3.4cm]{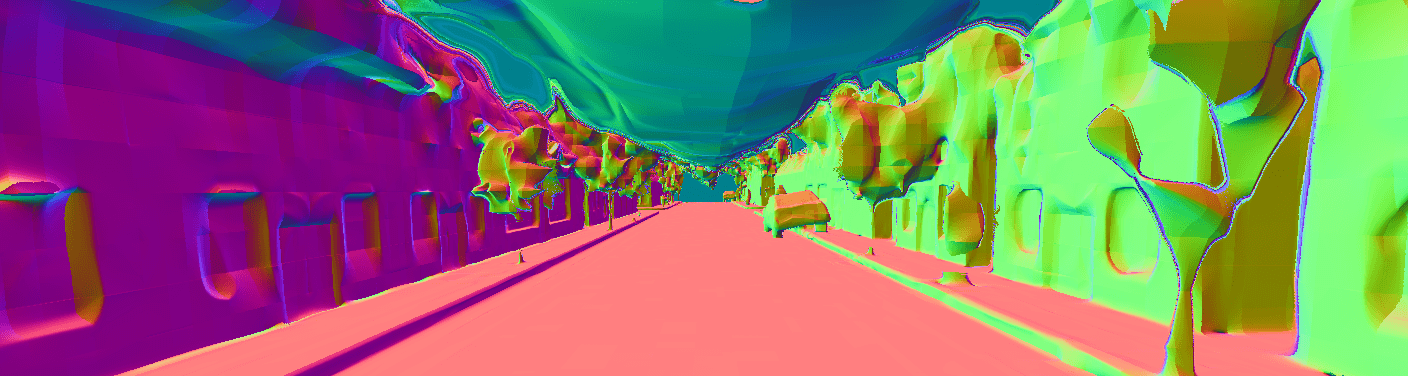}
    }\hspace{-3.5mm}
    \subfigure[Dense Grid]{
        \includegraphics[width=3.4cm]{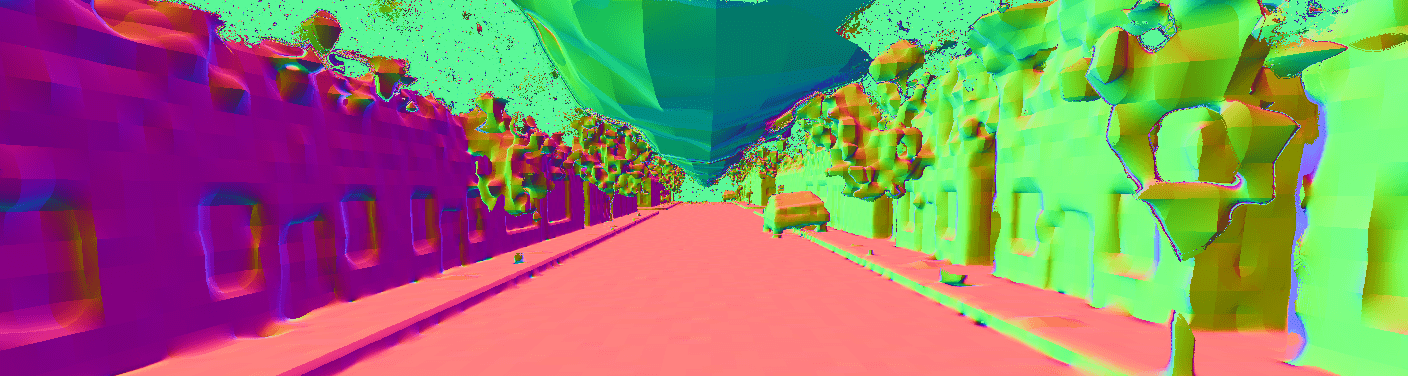}
    }\hspace{-3.5mm}
    \subfigure[MLP]{
        \includegraphics[width=3.4cm]{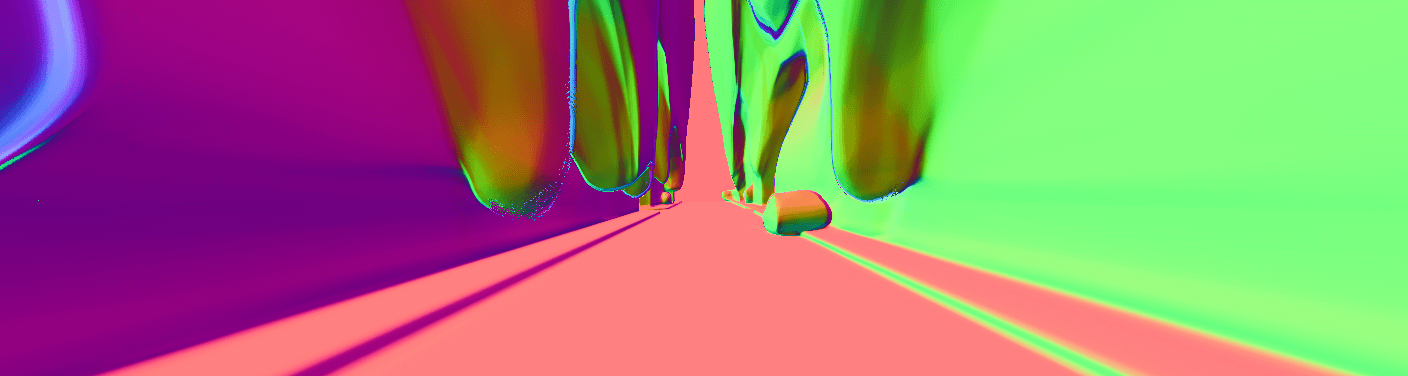}
    }
    \caption{Case comparison of different model architecture on the Maicity. The first row shows the reconstruction, whose details on pedestrians are highlighted in the white box in the lower left corner. The second row shows the rendered normal images for the specified viewpoint.}
    \vspace{-2mm}
\label{fig2}
\end{figure*}

\section{Experimental Results}

We validate the effectiveness of our approach on simulation scenes and real scenes. The impact of additional terms is also verified. We conduct comparison and performance analysis. In addition, we explore the effectiveness of using our atlas to enable global representations.

\subsection{Setup}
\textbf{Dataset:}  Our study involves experiments on two outdoor urban datasets: Maicity \cite{maicity} and KITTI-360 \cite{kitti360}, and two indoor datasets: Lobby dataset collected by ourselves and Hilti SLAM 2021 \cite{hilti}. Outdoor datasets were captured using the Velodyne HDL-64 LiDAR. While Maicity is a smaller synthetic dataset, it possesses ground truth mesh and high-fidelity LiDAR measurements, which make it suitable for evaluation tests. In contrast, KITTI-360 is larger in size and has noisy measurements, making it appropriate for both local and global evaluation tests. Besides, KITTI-360 also provides semantic cues and filters out dynamic objects, which enables us to evaluate semantic reconstruction. The setup and results for indoor datasets are shown in Appendix. 


\textbf{Metrics:}  We conduct a quantitative evaluation of the reconstruction quality of generated maps by comparing the densely sampled point cloud from observed surfaces to the reference point cloud. The reference point cloud for Maicity is obtained from a CAD model, while for the KITTI-360 dataset, we follow the evaluation criterion of other work \cite{vdbfusion}. We employ completeness, accuracy, Chamfer-L1 distance, recall, precision and F-score. The recall, precision and F-score are computed using a threshold of 5cm for Maicity and 10cm for KITTI-360, respectively. Additionally, we measure the training time and model memory for reconstruction, as well as the time taken for inference. 

\begin{table}[t]
    \centering
    \setlength{\tabcolsep}{4pt}
    \caption{Mapping quality using different model architectures}
    \resizebox{\linewidth}{!}{
    \begin{tabular}{llcccccl}
        \toprule
            \textbf{Data} &
            \textbf{Model} & \textbf{Comp.}$\downarrow$ & \textbf{Acc.}$\downarrow$ & \textbf{C-L1}$\downarrow$ & \textbf{Re.}$\uparrow$ & \textbf{Pre.}$\uparrow$ & \textbf{F1}$\uparrow$\\
        \midrule
            \multirow{5}*{Maicity} & MLP & 9.26 & 2.59 & 5.93 & 79.90 & 91.61 & 85.36 \\
            & DG & 1.15 & 1.49 & 1.32 & 97.19 & 97.07 & 97.13 \\
            & DFG+MLP & 1.11 & 1.39 & 1.25 & 97.43 & 97.65 & 97.54 \\
            & HFG+MLP & 1.07 & 1.40 & 1.23 & 97.88 & 97.87 & 97.88 \\
            & \textbf{Ours} & \textbf{0.95} & \textbf{1.33} & \textbf{1.14} & \textbf{98.87} & \textbf{97.71} & \textbf{98.28} \\
        \midrule
            \multirow{5}*{KITTI-360} & MLP & 13.83 & 5.40 & 9.61 & 60.87 & 86.34 & 71.40 \\
            & DG & 5.85 & 4.45 & 5.15 & 85.33 & 89.51 & 87.37 \\
            & DFG+MLP &  7.72  & 5.36 & 6.54 & 78.23 & 86.58 & 82.20 \\
            & HFG+MLP & 7.39 & 5.81 & 6.60 & 80.12 & 84.69 & 82.34\\
            & \textbf{Ours} & \textbf{4.06} & \textbf{3.39} & \textbf{3.82} & \textbf{93.86} & \textbf{95.23} & \textbf{94.54} \\
        \bottomrule
    \end{tabular}
    }
    \label{tab1}
    
\end{table}
\begin{table}[t]
    \centering
    \setlength{\tabcolsep}{4pt}
    \caption{\label{tab2}Memory and computation time using different model architecture}
    \resizebox{\linewidth}{!}{
    \begin{tabular}{llcccl}
        \toprule
            \textbf{Data} & \textbf{Model} & \textbf{Memory} $\downarrow$ & \textbf{Train Time} $\downarrow$ & \textbf{Render Time} $\downarrow$ \\
        \midrule
            \multirow{5}*{Maicity} & MLP & \textbf{1.32M} & 207.9min & 13.13s \\
             & DG & 2781.0M & 29.1min & \textbf{0.28s}&\\
             & DFG+MLP & 1755.4M & 51.8min & 1.01s \\
             & HFG+MLP & 160.5M & 45.6min & 0.86s \\
             & \textbf{Ours} & 64.3M & \textbf{11.3min} & 0.30s \\
        \midrule
            \multirow{5}*{KITTI-360} & MLP & \textbf{1.32M} & 378.6min & 15.49s \\
             & DG & 2781M & 57.8min & 0.28s \\
             & DFG+MLP & 1755.38M & 72.8min & 1.14s \\
             & HFG+MLP & 164.51M & 55.3min & 0.92s \\
             & \textbf{Ours} & 26.91M & \textbf{39.0min} & \textbf{0.26s} \\
        \bottomrule
    \end{tabular}
    }
    \vspace{-0.3cm}
\end{table}


\textbf{Implementation:} We employ a sequence of 80 consecutive poses and their measurements for training a volume. We fix poses of the overlapping part and optimize new poses jointly with the volume features and the small MLP. We use top 3 levels features in the Octree where the top level resolution depends on the size of the scene. The SDF branch shares a layer of 128 hidden units with the semantic branch, and the semantic branch is followed by another layer of 64 hidden units. We sample 1024$\times$80 rays per batch. On each ray, we have 12 samples using voxel-guided sampling and 12 samples using surface-guided sampling within the range of 25m. Thanks to the CUDA implementation, it takes on average $700ms$ for each mapping iteration which is $2\times$ faster than double back-propagation without CUDA. All experiments run on a single RTX-3090 GPU. 

\textbf{Baselines:} We adopt the Voxblox~\cite{voxblox} and 3D SIREN~\cite{shi2022city} methods as conventional and learning-based baseline methods. These two methods are shown to have ability to map large-scale urban scenes with LiDAR scans. The source code of Voxblox is accessible. We use the original parameters on KITTI. Regarding 3D SIREN, we have implemented the method in accordance with the original paper~\cite{shi2022city}.

\begin{table}[t]
    \centering
    \caption{\label{tab:smoothing_test}Mapping quality using model with and without the smoothness prior at different noise level on Maicity.}
    \resizebox{\linewidth}{!}{
    \begin{tabular}{cccccccl}
        \toprule
             \textbf{$\sigma_{noise}$} & \textbf{Model} & \textbf{Comp.}$\downarrow$ & \textbf{Acc.}$\downarrow$ & \textbf{C-L1}$\downarrow$ & \textbf{Re.}$\uparrow$ & \textbf{Pre.}$\uparrow$ & \textbf{F1}$\uparrow$\\
        \midrule
            \multirow{2}*{10} & w/o $p_h$ & 1.41 & 1.88 & 1.64 & \textbf{98.87} & 96.01 & 97.42\\
                & \textbf{Ours} & \textbf{1.29} & \textbf{1.70} & \textbf{1.50} & 98.86 & \textbf{96.52} & \textbf{97.67}\\
        \midrule
            \multirow{2}*{5} & w/o $p_h$ & 1.01 & 1.49 & 1.25 & \textbf{99.38} & \textbf{97.10} & \textbf{98.23}\\
                & \textbf{Ours} & \textbf{1.01} & \textbf{1.46} & \textbf{1.23} & 99.25 & 97.09& 98.16\\
                
        \bottomrule
    \end{tabular}
    }
    \vspace{-2mm}
\end{table}

\begin{figure}
   \centering
   \subfigure[Ours]{
       \includegraphics[width=4cm]{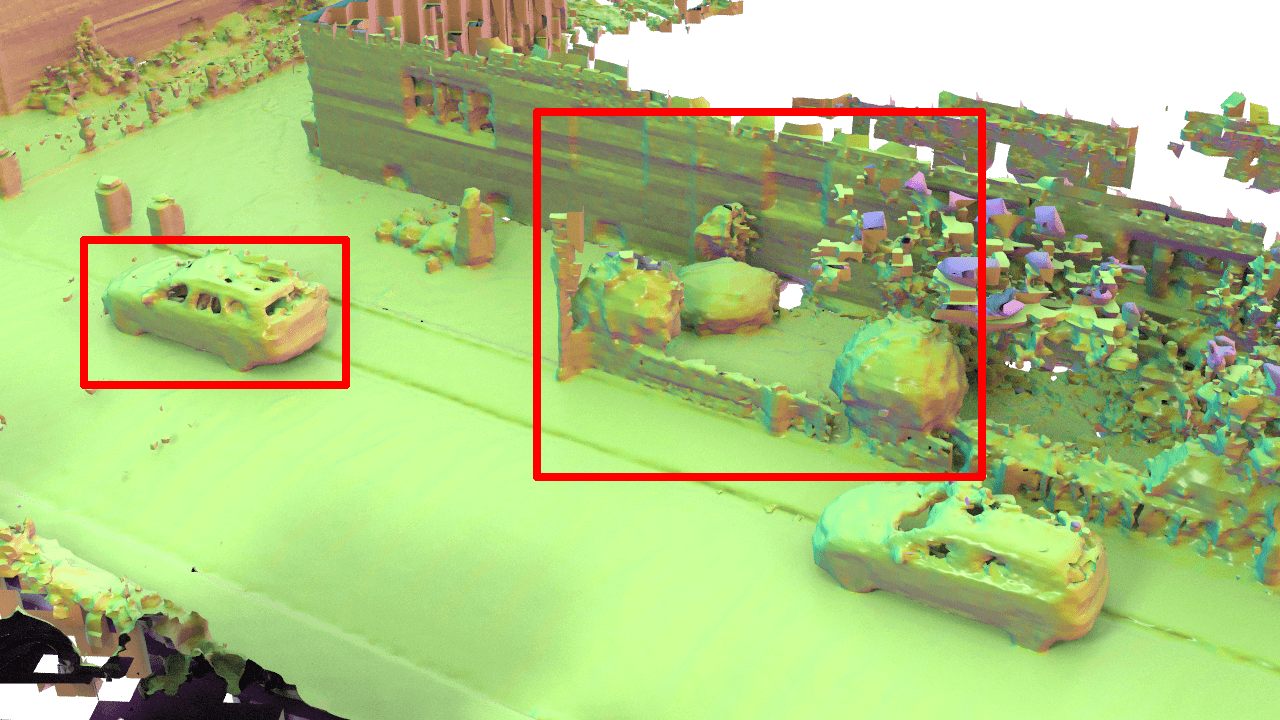} 
   }\hspace{-2mm}
   \subfigure[Ours w/o smoothness prior]{
       \includegraphics[width=4cm]{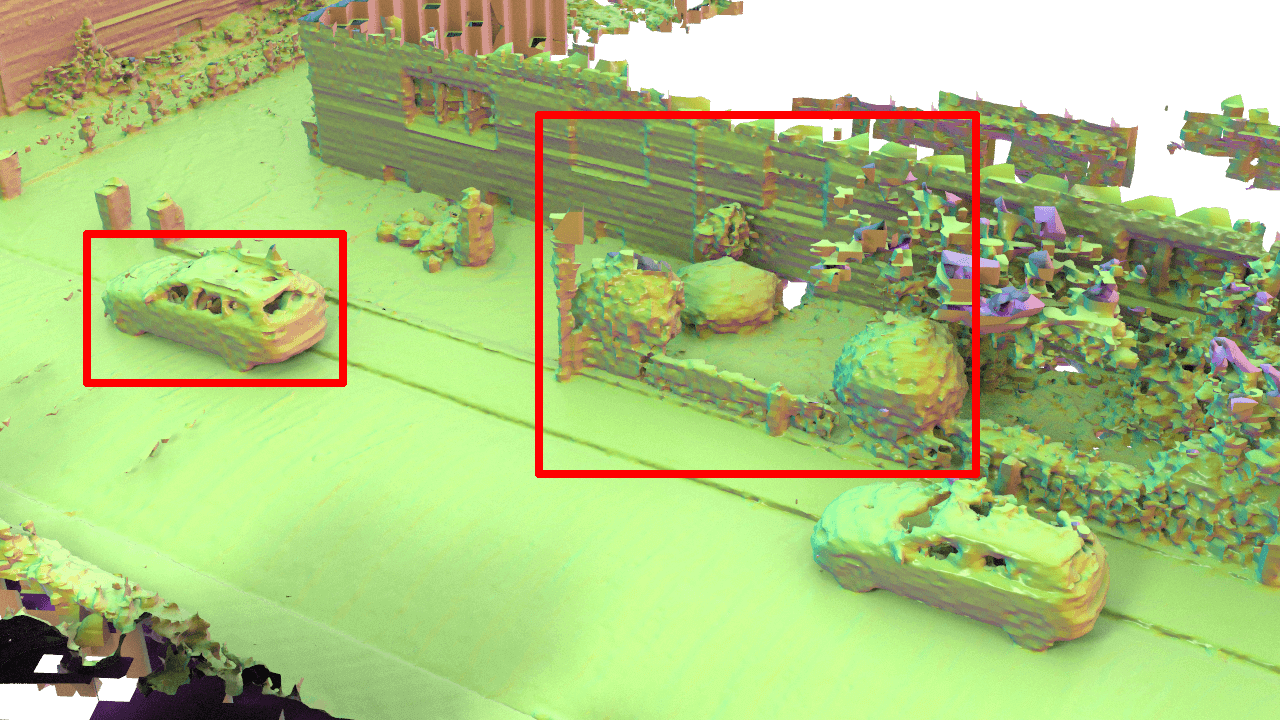} 
   }\hspace{-2mm}
   \caption{\label{spcase}Case comparison of our model with and without smoothness prior on KITTI-360. The red boxes highlight the difference in the surface details.}
   \vspace{-5mm}
\end{figure}

\subsection{Ablation Study}


\textbf{Effect of Feature Octree:}  We conduct a quantitative evaluation of 5 map representations on Maicity-01 and a 150m segment of Seq.00 of KITTI-360. The map representations that we evaluate include a large MLP~\cite{shi2022city}, Dense SDF grids (DG)~\cite{yu2021plenoxels}, a Dense Multi-Level Feature Grids with a small MLP (DFG+MLP)~\cite{wang2022gosurf,zhu2022nice}, and a Hash Indexed Multi-Level Feature Grids with a small MLP (HFG+MLP)~\cite{muller2022instantngp}. Notably, DG and DFG+MLP options are set at a resolution of $0.35m$ due to their high memory consumption. HFG+MLP has the same finest resolution of $0.088m$ as ours. Since the mapping quality of DFG+MLP and HFG+MLP can be improved by more levels, we set them to 4 and 5 levels respectively, both have more levels than ours.


Upon evaluation of both datasets, our method demonstrates the highest quality as shown in Tab.~\ref{tab1}. This can be attributed to its superior resolution compared to DG and DFG+MLP, as well as its less feature sharing in comparison to HFG+MLP. As shown in Tab.~\ref{tab2}, our method only requires a higher memory than MLP, while the latter exhibits weaker quality. In terms of training efficiency, our method outperforms all others. This is largely due to the adoption of a sparse Octree structure, which effectively reduces the number of features and improves sampling efficiency. For inference efficiency, ours remains competitive with the network-free DG, which is also brought by the Octree guided sampling strategy. In several cases shown in Fig.~\ref{fig2}, ours captures more intricate surface details, such as human legs.

\textbf{Effect of Smoothness Prior:} The effectiveness of the prior is evaluated on Maicity, as it provides ground truth data. With noiseless data, adding prior knowledge is counterproductive. In order to accurately reflect the influence of the smoothness prior, two levels of noise are manually added to the scans. Tab.~\ref{tab:smoothing_test} reports that as noise levels increase, the influence of smoothness becomes increasingly significant. This finding is reasonable given the reduction in measurement confidence under high noise conditions. Moreover, Fig.~\ref{spcase} illustrates that the smoothness prior effectively mitigates the noise in the reconstructed surface.


\begin{figure}[t]
    \flushleft
    \subfigure{
        \includegraphics[width=2.7cm]{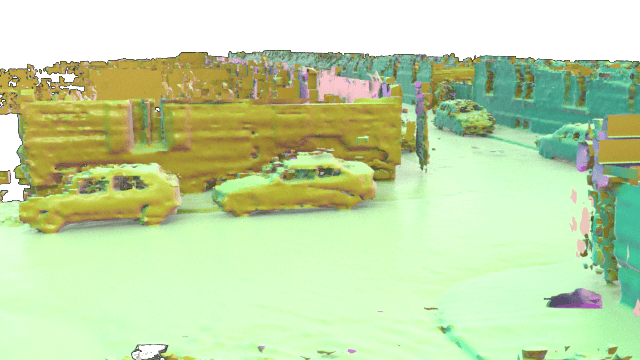} 
    }\hspace{-3.3mm}
    \subfigure{
        \includegraphics[width=2.7cm]{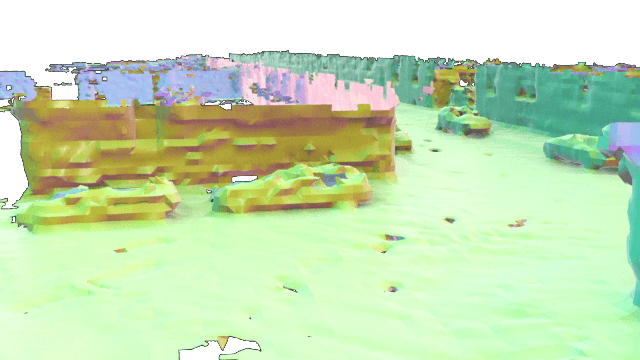} 
    }\hspace{-3.3mm}
    \subfigure{
        \includegraphics[width=2.7cm]{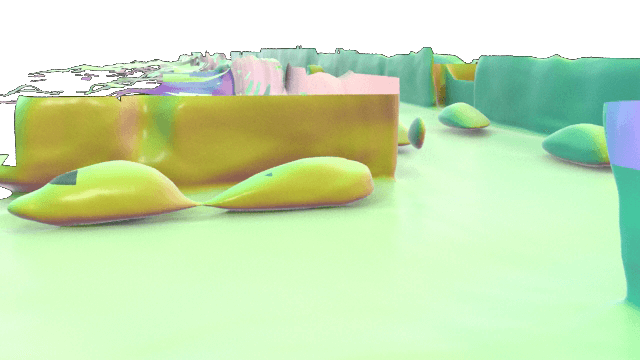} 
    }\hspace{-3.3mm}
    \subfigure{
        \includegraphics[width=2.7cm]{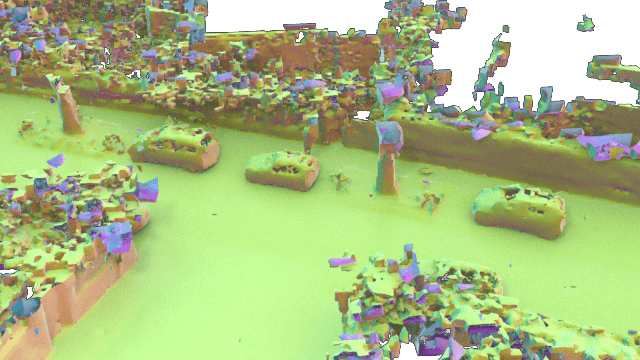} 
    }\hspace{-3.3mm}
    \subfigure{
        \includegraphics[width=2.7cm]{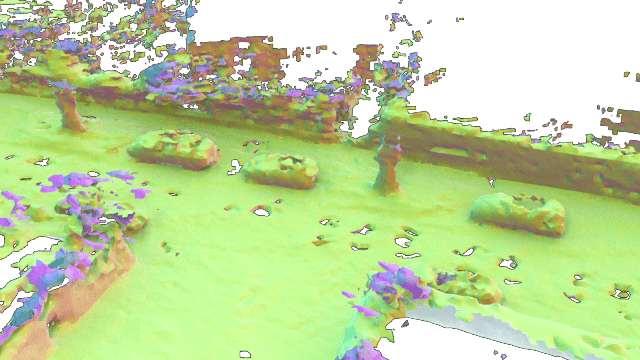} 
    }\hspace{-3.3mm}
    \subfigure{
        \includegraphics[width=2.7cm]{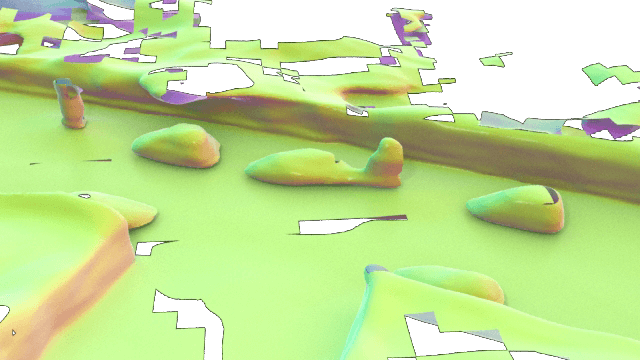} 
    }\hspace{-3.3mm}
    \setcounter{subfigure}{0}
    \subfigure[Ours]{
        \includegraphics[width=2.7cm]{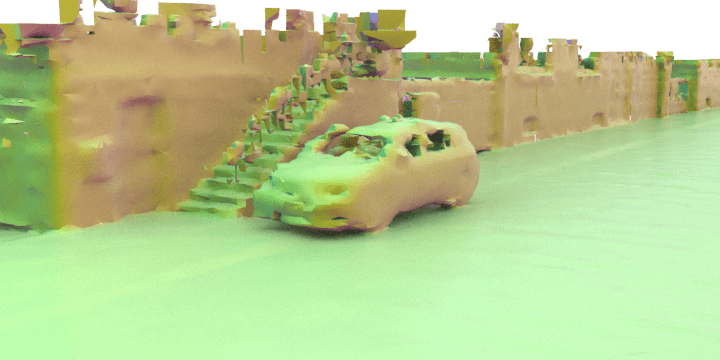} 
    }\hspace{-3.3mm}
    \subfigure[Voxblox]{
        \includegraphics[width=2.7cm]{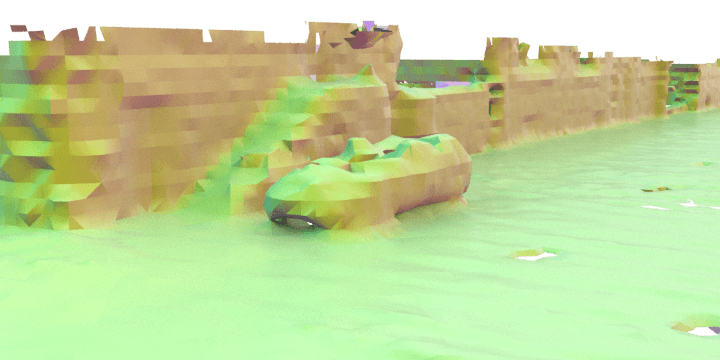} 
    }\hspace{-3.3mm}
    \subfigure[3D-SIREN]{
        \includegraphics[width=2.7cm]{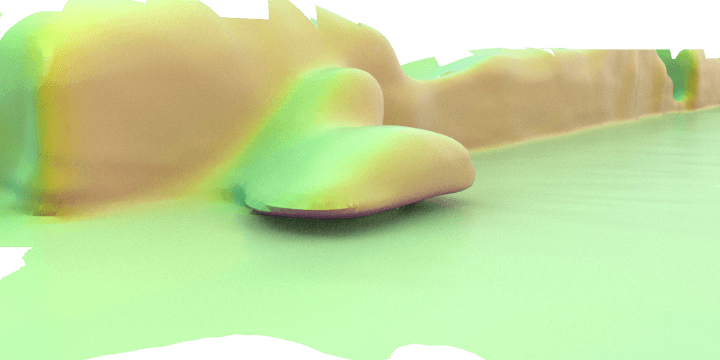} 
    }\hspace{-3.3mm}
    \caption{\label{qualitycompcase}Case comparison of different methods on the KITTI-360. Each row shows the results of a case area using different methods.}
    \vspace{-3mm}
\end{figure}

\begin{table}[t]
    \centering
    \caption{\label{tab:test}Mapping quality using different methods on KITTI-360}
    \resizebox{\linewidth}{!}{
    \begin{tabular}{lccccccl}
        \toprule
             \makebox[0.01\textwidth][c]{\textbf{Seq.}} & \textbf{Method} & \textbf{Comp.}$\downarrow$ & \textbf{Acc.}$\downarrow$ & \textbf{C-L1}$\downarrow$ & \textbf{Re.}$\uparrow$ & \textbf{Pre.}$\uparrow$ & \textbf{F1}$\uparrow$\\
        \midrule
            \makebox[0.01\textwidth][c]{\multirow{3}*{00}} & 3D-SIREN & 7.61 & 8.99 & 8.30 & 74.69 & 78.64 & 76.62 \\
            & Voxblox & 11.65 & 4.59 & 8.12 & 68.66 & 90.42 & 78.05 \\
                & \textbf{Ours} & \textbf{5.07} & \textbf{3.79} & \textbf{4.43} & \textbf{89.60} & \textbf{94.85} & \textbf{92.15} \\
        \midrule
            \makebox[0.01\textwidth][c]{\multirow{3}*{02}} & 3D-SIREN & 12.37 & 8.34 & 10.35 & 74.56 & 69.37 & 71.87 \\
            & Voxblox & 8.70 & 5.21 & 6.95 & 76.81 & 87.28 & 81.71 \\
                & \textbf{Ours} & \textbf{4.76} & \textbf{3.99} & \textbf{4.37} & \textbf{90.81} & \textbf{94.07} & \textbf{92.41} \\
        \midrule
            \makebox[0.01\textwidth][c]{\multirow{3}*{04}} & 3D-SIREN & 7.47 & 8.71 & 8.09 & 74.99 & 82.78 & 78.69\\
            & Voxblox & 8.83 & 4.80 & 6.81 & 77.71 & 90.55 & 83.64 \\
                & \textbf{Ours} & \textbf{4.50} & \textbf{3.57} & \textbf{4.04} & \textbf{92.47} & \textbf{95.33} & \textbf{93.88} \\
        \bottomrule
    \end{tabular}
    }
    \vspace{-4mm}
\end{table}

\textbf{Effect of Semantic Measurement:} We evaluate the effectiveness of semantic measurements on KITTI-360. The results are slightly surprising that the semantics can marginally improve the mapping quality in addition to purely assigning labels. The C-L1 and F-score are improved from $5.70cm$ to $5.58cm$, and $87.66\%$ to $88.80\%$. This result may be attributed to the advantages of multi-task learning. Two cases of the semantic reconstruction are shown in Fig.~\ref{hole}.



\subsection{Comparative Study}
\begin{figure}[t]
    \flushleft
    \subfigure{
        \includegraphics[width=4.1cm]{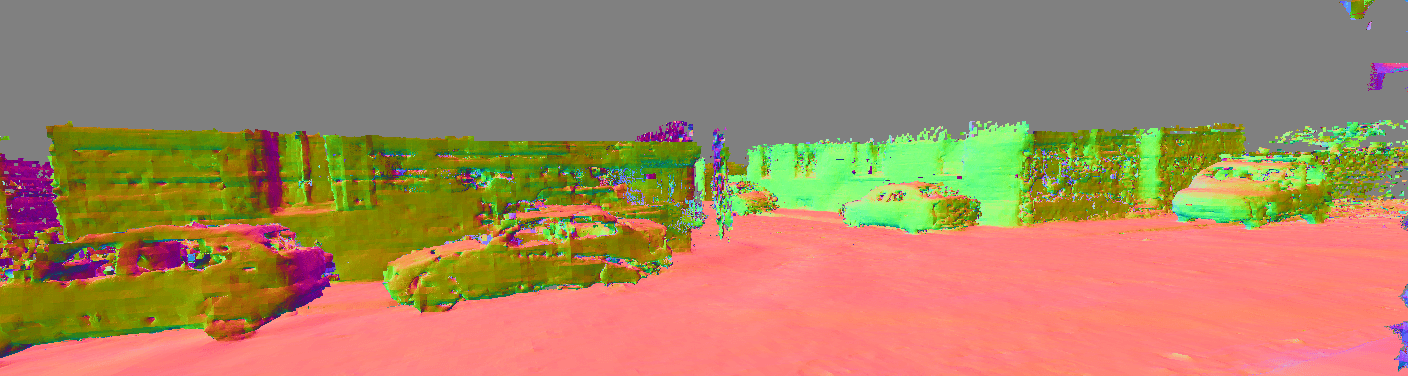} 
    }\hspace{-2mm}
    \subfigure{
        \includegraphics[width=4.1cm]{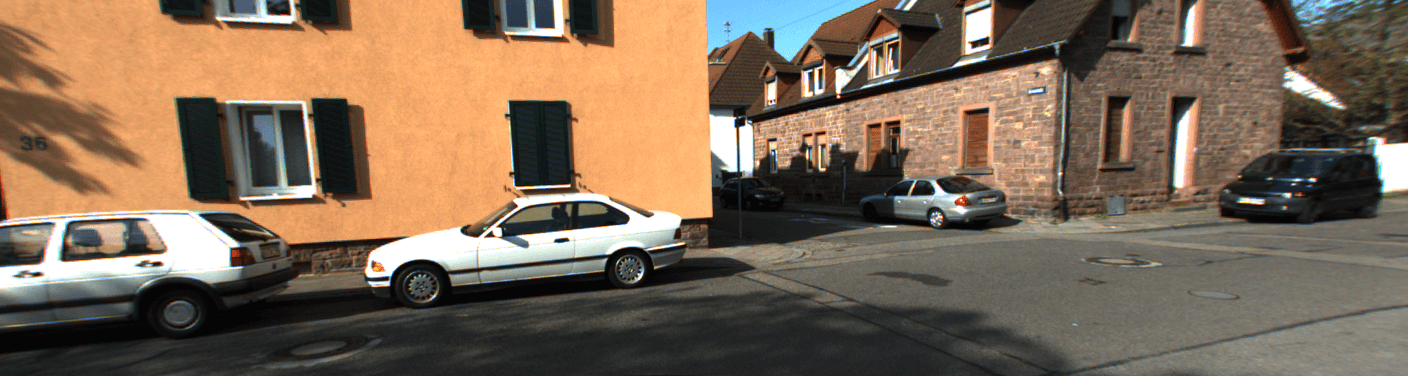} 
    }\hspace{-2mm}
    \subfigure{
        \includegraphics[width=4.1cm]{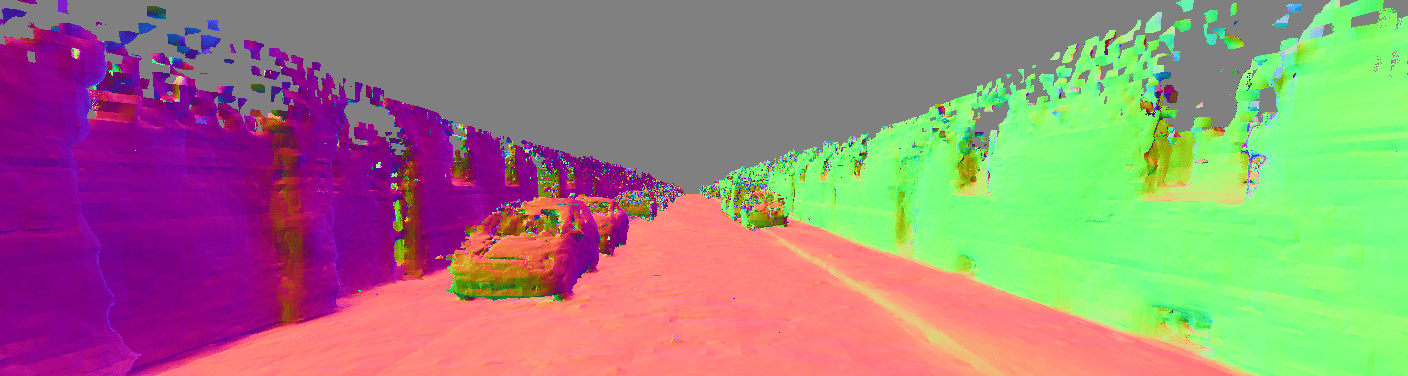} 
    }\hspace{-2mm}
    \subfigure{
        \includegraphics[width=4.1cm]{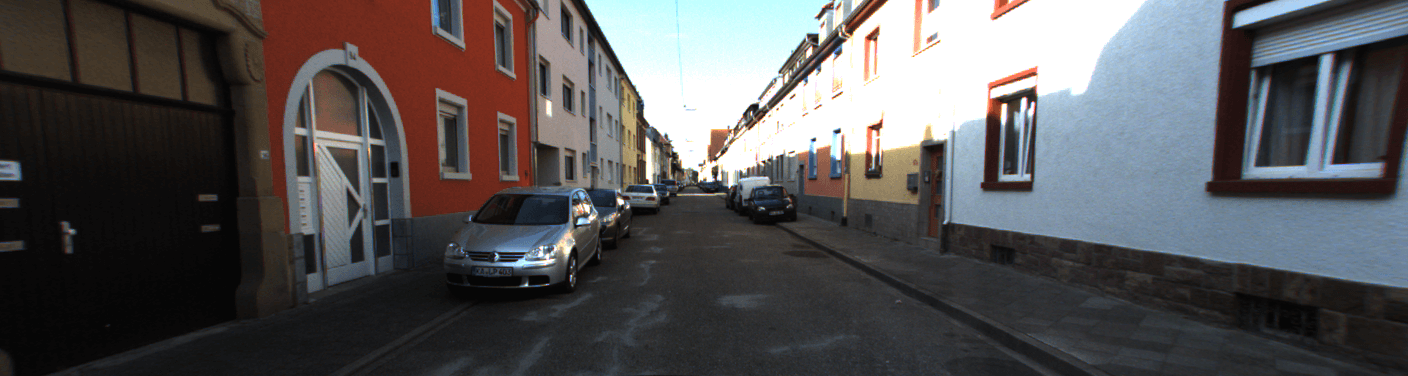} 
    }\hspace{-2mm}
   \subfigure{
        \includegraphics[width=4.1cm]{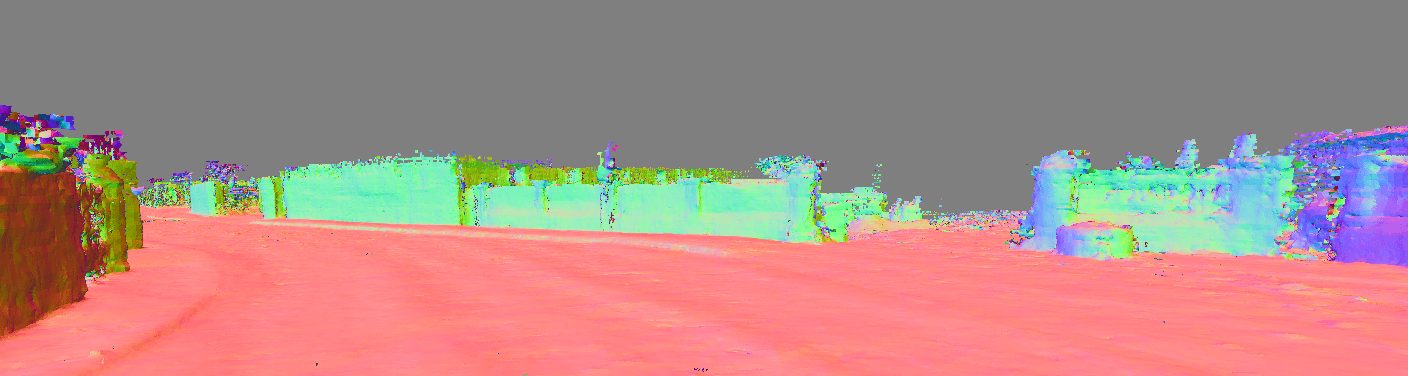} 
    }\hspace{-2mm}
    \subfigure{
        \includegraphics[width=4.1cm]{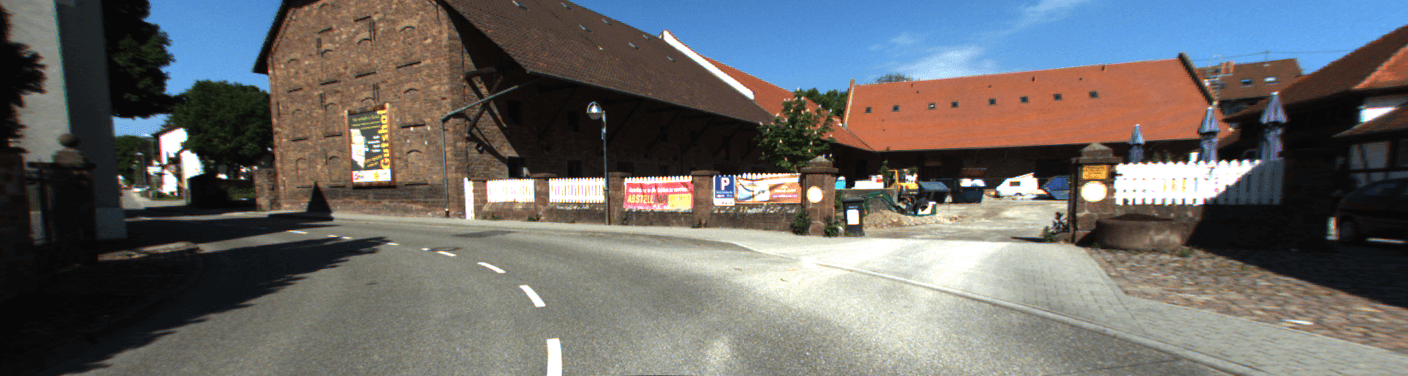} 
    }\hspace{-2mm}
    \subfigure{
        \includegraphics[width=4.1cm]{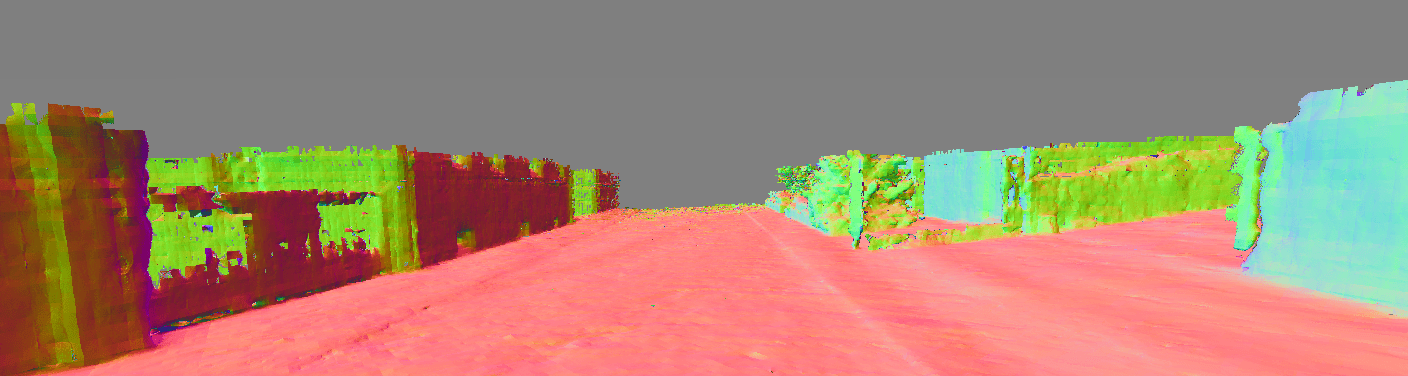} 
    }\hspace{-2mm}
    \subfigure{
        \includegraphics[width=4.1cm]{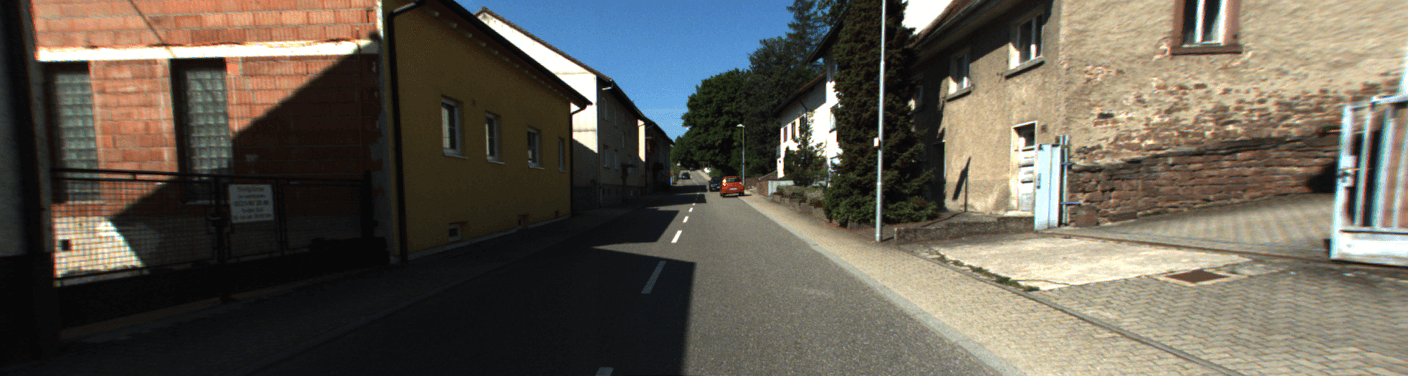} 
    }\hspace{-2mm}
    \caption{\label{rendercase}Cases of rendered normal images (left column) and corresponding real images (right column) using our method on KITTI-360.}
\end{figure}

Our approach is compared against two existing methods, 3D-SIREN~\cite{shi2022city} and Voxblox~\cite{voxblox}, on three sequences from the KITTI-360 dataset for testing: scans 6330$\sim$6530 of Seq.00, scans 4485$\sim$4685 of Seq.02, and scans 5940$\sim$6140 of Seq.04. The results presented in Tab.~\ref{tab:test} demonstrate that our method produces more complete and accurate results than others. Specifically, the volume feature in our approach enhances fitting capacity compared to 3D-SIREN, while the end-to-end MAP optimization outperforms the updating rule of Voxblox. Fig.~\ref{qualitycompcase} displays examples where our method preserves details such as car windows and stairs. Fig.~\ref{rendercase} illustrates examples of rendered normal images, showing consistent surface structures with the ones in real images.

\begin{figure}[t]
   \flushleft
   \subfigure{
       \includegraphics[width=4.2cm]{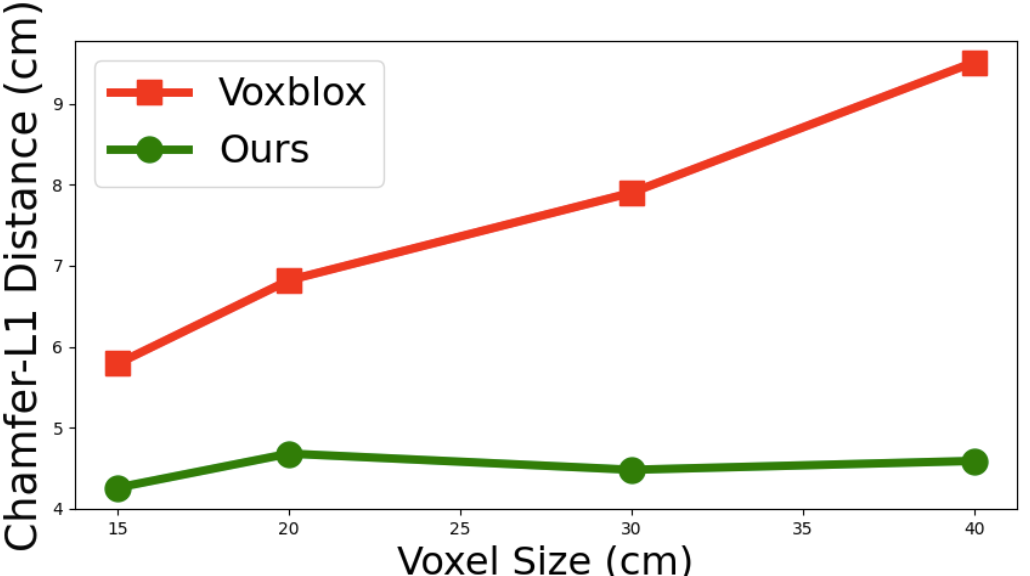} 
   }\hspace{-4mm}
   \subfigure{
       \includegraphics[width=4.2cm]{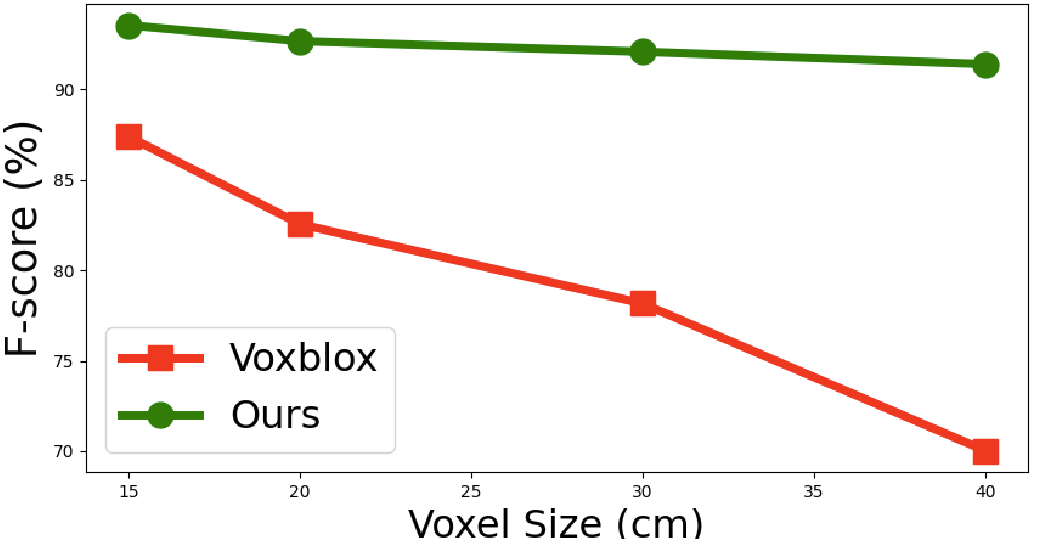} 
   }\hspace{-4mm}
   \caption{\label{voxelsizecomp} Mapping quality with respect to voxel size on KITTI-360.}
   \vspace{-3mm}
\end{figure}

We further assess the reconstruction quality with different voxel sizes on Seq.02 of KITTI-360. As depicted in Fig.~\ref{voxelsizecomp}, our approach outperforms Voxblox across all voxel size configurations. Furthermore, our method is less prone to degradation of mapping quality as voxel size increases. This phenomenon can be attributed to the employment of neural features, which possess the potential for super-resolution.


\subsection{Case Study on Incremental Mapping} 
\begin{figure}[t]
    \flushleft
    \subfigure[Ours]{
        \includegraphics[width=4cm]{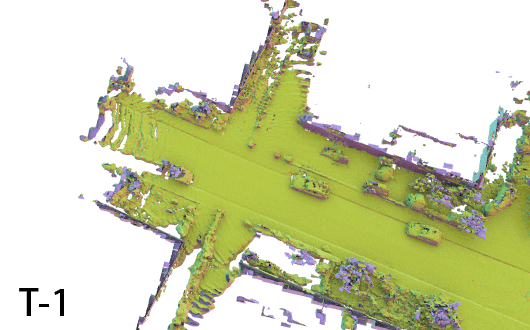} \label{1}
    }\hspace{-3mm}
    \subfigure[Ours]{
        \includegraphics[width=4cm]{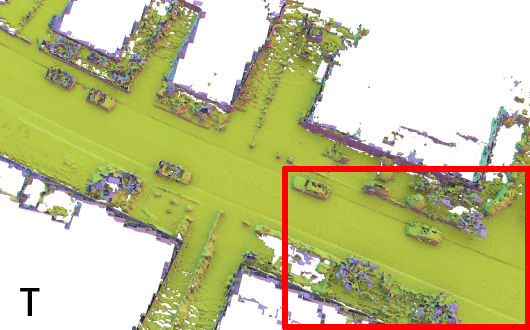} \label{2}
    }\hspace{-3mm}
    \subfigure[HFG+MLP]{
        \includegraphics[width=4cm]{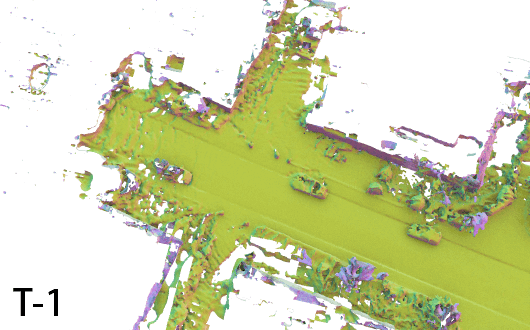} \label{3} 
    }\hspace{-3mm}
    \subfigure[HFG+MLP]{
        \includegraphics[width=4cm]{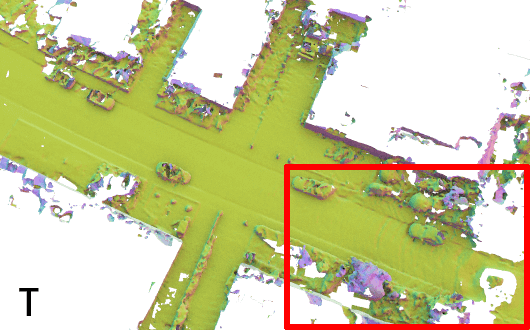} \label{4} 
    }\hspace{-3mm}
    \caption{\label{continual}Case comparison of different methods on the KITTI-360. Each row shows the results of incremental mapping from previous time to current time. The red boxes highlight the change in the previous map caused by the catastrophe forgetting.}
    \vspace{-1mm}
\end{figure}

In incremental mapping, the accumulation of new measurements inevitably limits the proportion of past measurements that can be retained due to constraints in memory and computational resources. This phenomenon leads to the catastrophic forgetting of previously acquired mapping information within the network. However, NF-Atlas allows each volume to focus solely on local area, thus effectively avoiding catastrophic forgetting. Fig.~\ref{continual} supports our claim, while single model methods such as HFG+MLP suffer from a noticeable decline in surface detail retention of past maps.


By integrating semantic measurements, the outcome of an incremental semantic mapping over a more than $1km$ trajectory in KITTI-360 is depicted in Fig.~\ref{hole}, which verifies that our approach is capable of large scale LiDAR mapping. 


\subsection{Case Study on Loop Closure}
\begin{figure}[t]
    \centering
    \subfigure[Before loop closure]{
        \includegraphics[width=4cm]{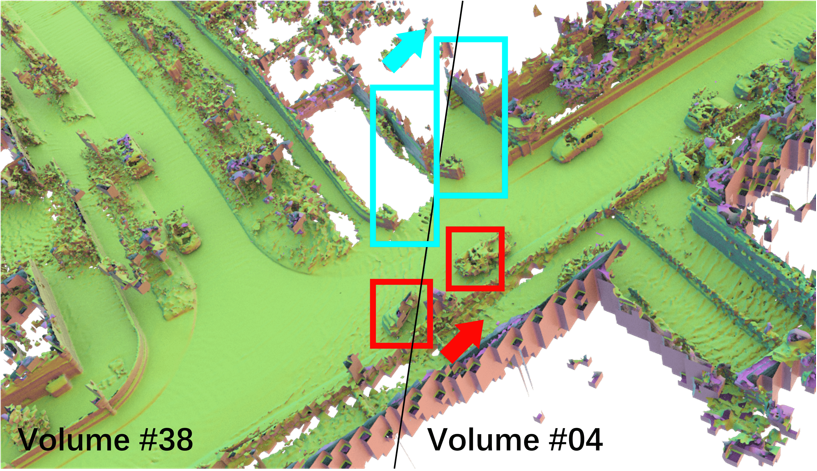} \label{1}
    }\hspace{-2mm}
    \subfigure[After loop closure]{
        \includegraphics[width=4cm]{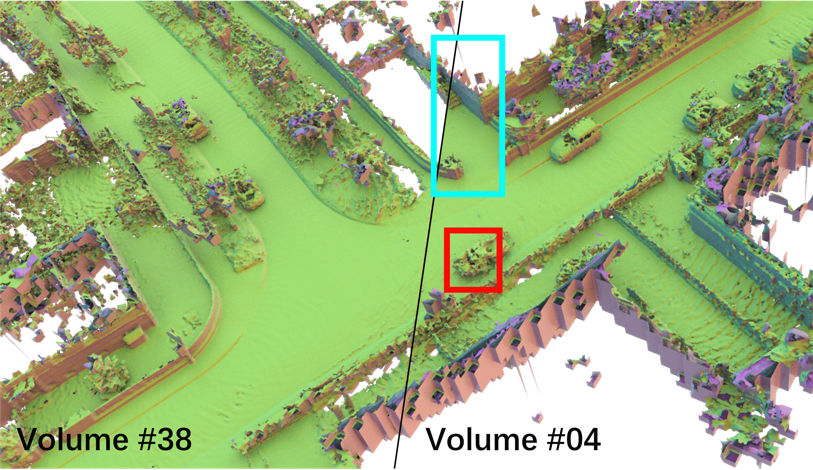} \label{2}
    }\hspace{-2mm}
    \caption{\label{loopmap}Case of mapping reuslt when a loop closure occurs using our method on KITTI-360. The red box and the blue box highlight that the objects are aligned after the loop closure. The black line is the boundary between the two volumes stitched in the global map.}
    \vspace{-4mm}
\end{figure}

Upon the occurrence of loop closure, the trajectory updates through pose graph optimization. By representing the map as elastic neural feature fields, our method avoids the remapping along the updated trajectory. In Fig.~\ref{loopmap}, the results before and after the loop closure are demonstrated. By only updating the origins of submaps, the global map remains consistent, yielding a substantial reduction in computation.

\section{Conclusion}

We propose multi-volume neural feature fields, NF-Atlas, for large-scale LiDAR mapping which combines the advantage of the neural implicit representation and the pose graph. In local, we state the mapping as a MAP problem, allowing for end-to-end optimization. The sparsity of feature Octree also improves efficiency and memory usage. In global, we fix the volumes to the pose graph nodes, allowing for only origin updating when loop closure occurs. The volume-by-volume process enables incremental mapping. On both simulation and real-world datasets, NF-Atlas is shown to be a competitive method. In the future, we set to deal with the dynamic objects in the environment.

\bibliographystyle{IEEEtran}
\bibliography{IEEEabrv,bibliography}

\section*{Appendix}



\subsection{Optimization of MAP}

By taking the logarithm, the MAP problem: 
\begin{equation}
\tilde{\theta},\tilde{\Psi},\tilde{\bo},\tilde{\bd} = \arg\max \prod {p_r \cdot p_s \cdot p_c \cdot p_e \cdot p_h} \tag {1}
\end{equation}
can be converted into a loss function:
\begin{equation}
\begin{aligned}
    &\tilde{\theta},\tilde{\Psi},\tilde{\bo},\tilde{\bd} = \arg\max \log (\prod {p_r \cdot p_s \cdot p_c \cdot p_e \cdot p_h})  \\
    &= \arg\max - \sum (\lambda_r L_r + \lambda_s L_s + \lambda_c L_c + \lambda_e L_e +\lambda_h L_h) \\
    &= \arg\min \sum (\lambda_r L_r + \lambda_s L_s + \lambda_c L_c + \lambda_e L_e +\lambda_h L_h)
\end{aligned} \tag {2}
\end{equation}
where $L_r$ is range loss:
\begin{equation}
    L_r = \frac{(\hat{r}-r(\bo,\bd))^2}{2\sigma_r^2} \tag {3}
\end{equation}
$L_s$ is SDF loss:
\begin{equation}
    L_s = \begin{cases}
     \hfil \frac{| b -s(\bp)|}{\lambda} & |b|\leq \tau\\ 
     \hfil \max(0, e^{-\beta s(\bp)}-1, s(\bp) - b ) & o.w. 
    \end{cases} \tag {4}
\end{equation}
$L_c$ is semantic loss:
\begin{equation}
     L_c = \sum\limits^C\limits_{c=1}{l_c(c) \log \hat{l}_c(c)} \tag {5}
\end{equation}
where $\hat{l}_c(c)$ is the multi-class semantic probability at class $c$ of the ground truth map.
$L_e$ is Eikonal loss:
\begin{equation}
     L_e = \frac{(1-\|\nabla_{\bp} s(\bp)\|)^2 }{2\sigma_e^2}) \tag {6}
\end{equation}
$L_h$ is smoothness loss:
\begin{equation}    
     L_h = \frac{\| \nabla_{\bp} s(\bp) - \nabla_{\bp} s(\bp+\Delta \bp) \|^2}{2\sigma_h^2} \tag {7}
\end{equation}

That is to convert solving the maximal posterior problem to minimize losses. With this loss function, we use the Adam optimizer based on gradient descent to optimize the iterative solution, which is similar to the training stage of standard differential rendering-based reconstruction methods\cite{wang2021neus,sun2022neuralinthewild,wang2022gosurf,yu2021plenoxels}.

The main reason to formulate a probabilistic mapping problem is to compare it with the classic probabilistic mapping in robotics. Note that the ray based likelihood is built upon multiple voxels. The classic way of updating SDF values\cite{voxblox} follows a strong assumption that each voxel in the posterior is independent, which does not follow the Bayesian rule. While in our method, voxels are updated exactly following the Bayesian rule, thus correlated. We consider that this reason mainly explains the more completed and accurate mapping result over the traditional one. In addition, this formulation also allows for more prior and likelihood factors, which may not be easy in the classic updating.

The first-order derivative and inverse second-order derivative of trilinear interpolation of the octree-based grid are implemented on CUDA operators. The details of the derivation of the gradient for trilinear interpolation of the octree grid are as follows:

\textbf{Octree-based Grid:}
As described in Section III-A, given a query point $\bp$, its feature is acquired by summing the tri-linearly interpolated features in top $K$ levels as: 
\begin{equation}
    \Psi(\bp)= \sum_{i=0}^{K-1} triInterp(\psi_{i,j\in \mathcal{N}(\bp)}) \tag{8}
\end{equation}
where $\mathcal{N}(\bp)$ is the set of the eight nearest Octree nodes i.e. corners$[(\lfloor x\rfloor ,\lfloor y\rfloor,\lfloor z\rfloor), (\lfloor x\rfloor ,\lfloor y\rfloor ,\lceil z\rceil), ..., (\lceil x\rceil ,\lceil y\rceil ,\lceil z\rceil)]^T$ of a cube containing $\bp=(x,y,z)$. In each octree level, 
 $\lfloor x\rfloor$ and $\lceil x\rceil$ are the largest integer less than $x$ and the smallest integer greater than $x$, within the range of the voxel of the current octree hierarchy in which $x$ is located.

\textbf{First-order Derivative:}
The trilinear interpolation $triInterp$ is calculated by first interpolating along the z-axis:

\begin{equation}
\begin{aligned}
    \displaystyle i_{1} &= \Psi[\lfloor x\rfloor ,\lfloor y\rfloor ,\lfloor z\rfloor ]\times (1-z_{d})+\Psi[\lfloor x\rfloor ,\lfloor y\rfloor ,\lceil z\rceil ]\times z_{d} \\
    \displaystyle i_{2} &= \Psi[\lfloor x\rfloor ,\lceil y\rceil ,\lfloor z\rfloor ]\times (1-z_{d})+\Psi[\lfloor x\rfloor ,\lceil y\rceil ,\lceil z\rceil ]\times z_{d} \\
    \displaystyle j_{1} &= \Psi[\lceil x\rceil ,\lfloor y\rfloor ,\lfloor z\rfloor ]\times (1-z_{d})+\Psi[\lceil x\rceil ,\lfloor y\rfloor ,\lceil z\rceil ]\times z_{d} \\
    \displaystyle j_{2} &= \Psi[\lceil x\rceil ,\lceil y\rceil ,\lfloor z\rfloor ]\times (1-z_{d})+\Psi[\lceil x\rceil ,\lceil y\rceil ,\lceil z\rceil ]\times z_{d}. 
\end{aligned} \tag{9}
\end{equation}

where $\displaystyle x_{d}=x-\lfloor x\rfloor$. Then, interpolating along the y-axis, we get:
\begin{equation}
\begin{aligned}
\displaystyle w_{1}&=i_{1}(1-y_{d})+i_{2}y_{d} \\
\displaystyle w_{2}&=j_{1}(1-y_{d})+j_{2}y_{d}
\end{aligned} \tag{10}
\end{equation}

Finally, interpolation along the x-axis yields to give the predicted value of the point:
\begin{equation}
\begin{aligned}
\displaystyle \Psi(\bp)&=w_{1}(1-x_{d})+w_{2}x_{d} \\
&=  \psi_{j\in \mathcal{N}(\bp)}^T w(\bp)
\end{aligned} \tag{11}
\end{equation}

where $w(\bp)$ is the interpolation coefficient vector:
\begin{equation}
w(\bp) =
\left[
    \begin{array}{c}
        (1-x)(1-y)(1-z) \\
        x(1-y)(1-z) \\
        (1-x)y(1-z) \\
        xy(1-z) \\
        (1-x)(1-y)z \\
        x(1-y)z \\
        (1-x)yz \\
        xyz
    \end{array}
\right]
 \tag{12}
\end{equation}

The Jacobian of $\Psi$ w.r.t. $\bp$ is given by:
\begin{equation}
\frac{\partial \Psi}{\partial \bp} = \psi_{j\in \mathcal{N}(\bp)}^T \frac{\partial w}{\partial \bp} 
\tag{13}
\end{equation}

where $\frac{\partial w}{\partial \bp}$ is the Jacobian of the interpolation coefficient vector w.r.t. $\bp$:
\begin{equation}
\frac{\partial w}{\partial \bp}  = 
\left[
    \setlength{\arraycolsep}{1.2pt}
    \begin{array}{ccc}
        -(1-y)(1-z) & -(1-x)(1-z) & -(1-x)(1-y)\\
        (1-y)(1-z) & -x(1-z) & -x(1-y)\\
        -y(1-z) & (1-x)(1-z) & -(1-x)y\\
        y(1-z) & x(1-z) & -xy\\
        -(1-y)z & -(1-x)z & (1-x)(1-y)\\
        (1-y)z & -xz & x(1-y)\\
        -yz & (1-x)z & (1-x)y\\
        yz & xz & xy
    \end{array}
\right] \tag{14}
\end{equation}
   
\textbf{Second-order Derivative:}
The second-order Derivative $\frac{\partial^2 \Psi}{\partial \bp^2}$ is given by:
\begin{equation}
\begin{aligned}
    \frac{\partial^2 \Psi}{\partial \bp^2} &= \frac{\partial }{\partial \bp} (\psi_{j\in \mathcal{N}(\bp)}^T \frac{\partial w}{\partial \bp} ) \\
    &= \psi_{j\in \mathcal{N}(\bp)}^T \frac{\partial^2 w}{\partial \bp^2} \\
\end{aligned} \tag{15}
\end{equation}
which is a $3\times3$ matrix. We define the result as:
\begin{equation}
\frac{\partial^2 \Psi}{\partial \bp^2} = 
\left[
    \setlength{\arraycolsep}{1.2pt}
    \begin{array}{ccc}
        v_{00} & v_{01} & v_{02}\\
        v_{10} & v_{11} & v_{12}\\
        v_{20} & v_{21} & v_{22}
    \end{array}
\right]  \tag{15}
\end{equation}
where 
\begin{equation}
\begin{aligned}
    v_{00} &= v_{11} = v_{22} = 0 \\
    v_{01} &= v_{10} = (1-z)(\psi_0 - \psi_1 - \psi_2 + \psi_3)\\ &+ z(\psi_4 - \psi_5 -\psi_6 + \psi_7) \\
    v_{02} &= v_{20} = (1-y)(\psi_0 - \psi_1 - \psi_4 + \psi_5)\\ &+ y(\psi_2 - \psi_3 -\psi_6 + \psi_7) \\
    v_{12} &= v_{21} = (1-x)(\psi_0 - \psi_2 - \psi_4 + \psi_6)\\ &+ x(\psi_1 - \psi_3 -\psi_5 + \psi_7) \\
\end{aligned}
\end{equation}

\subsection{Results on Indoor Datasets}
In this section, we provide a quantitative analysis of the reconstructed comparative experiments of two indoor datasets. Indoor data is more cluttered with objects, making it ideal for benchmarking neural fields. 

\textbf{Dataset:} 
We additionally do a comparison of the Voxblox reconstruction with our method on two indoor datasets which are captured using the OS0-64 LiDAR. One of them is recorded by ourselves using a mobile robot in the lobby of the university building. The other is the Hilti SLAM dataset 2021\cite{hilti}, recorded at the Hilti office using handheld LiDAR. 

\textbf{Setting:} 
Indoor scene experiments involving more cluttered objects require different volume sizes and several parameters for constraints. We use a $5$ cm leaf node resolution and a smaller perturbation vector in smoothness prior to better reconstruction performance. The poses are initialized by a LiDAR SLAM method. We follow the evaluation criterion and metrics of KITTI-360 mentioned in the main text. The recall, precision and F-score are computed using a threshold of 5cm for two indoor datasets.

\begin{figure}[H]
\renewcommand{\thefigure}{1}
\centering
\subfigure[Dataset]{
\begin{minipage}[b]{0.26\linewidth}
\includegraphics[height=1.8cm]{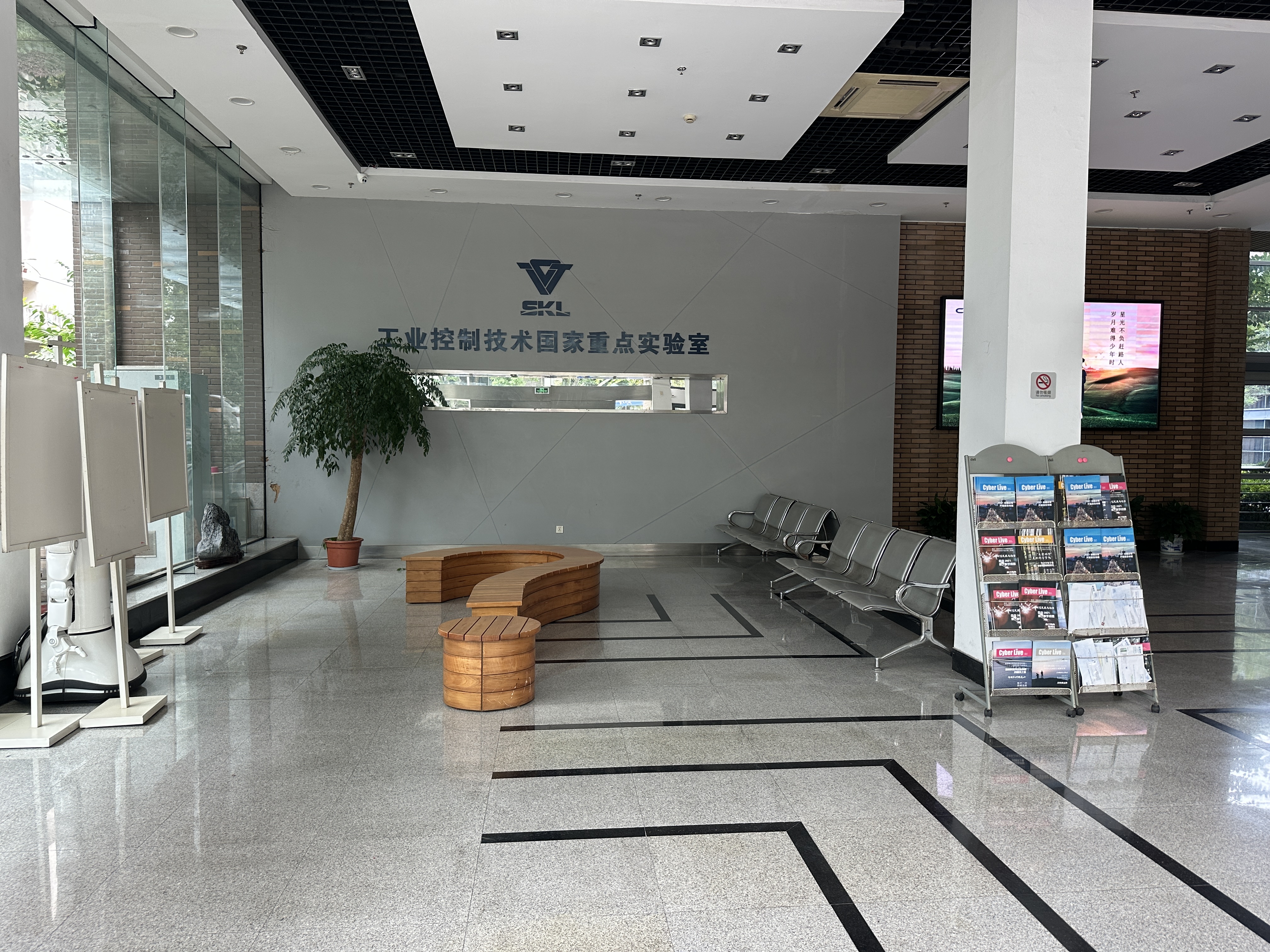}\vspace{4pt}
\includegraphics[height=1.8cm]{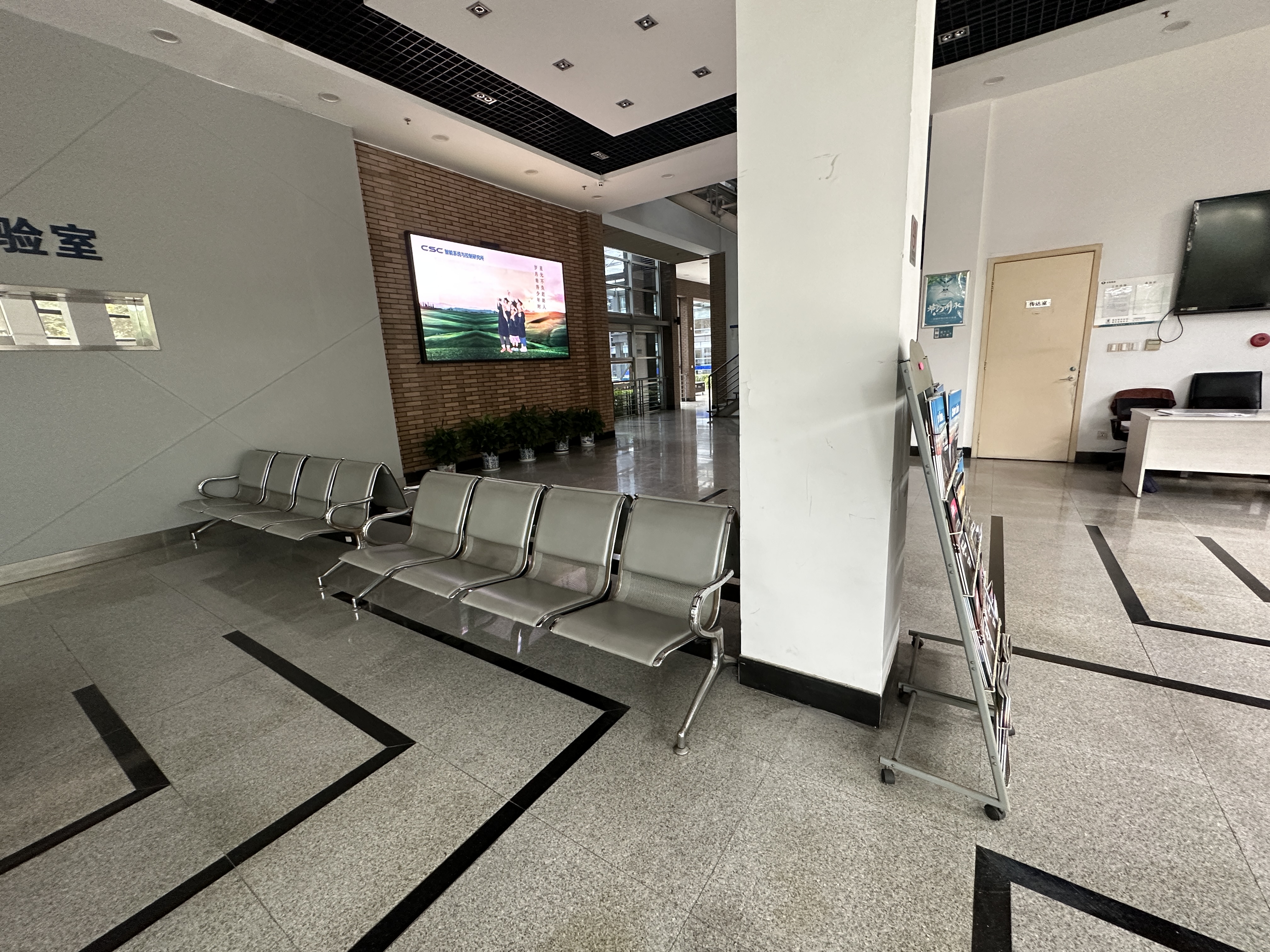}
\end{minipage}}
\subfigure[Ours]{
\begin{minipage}[b]{0.33\linewidth}
\includegraphics[height=1.8cm]{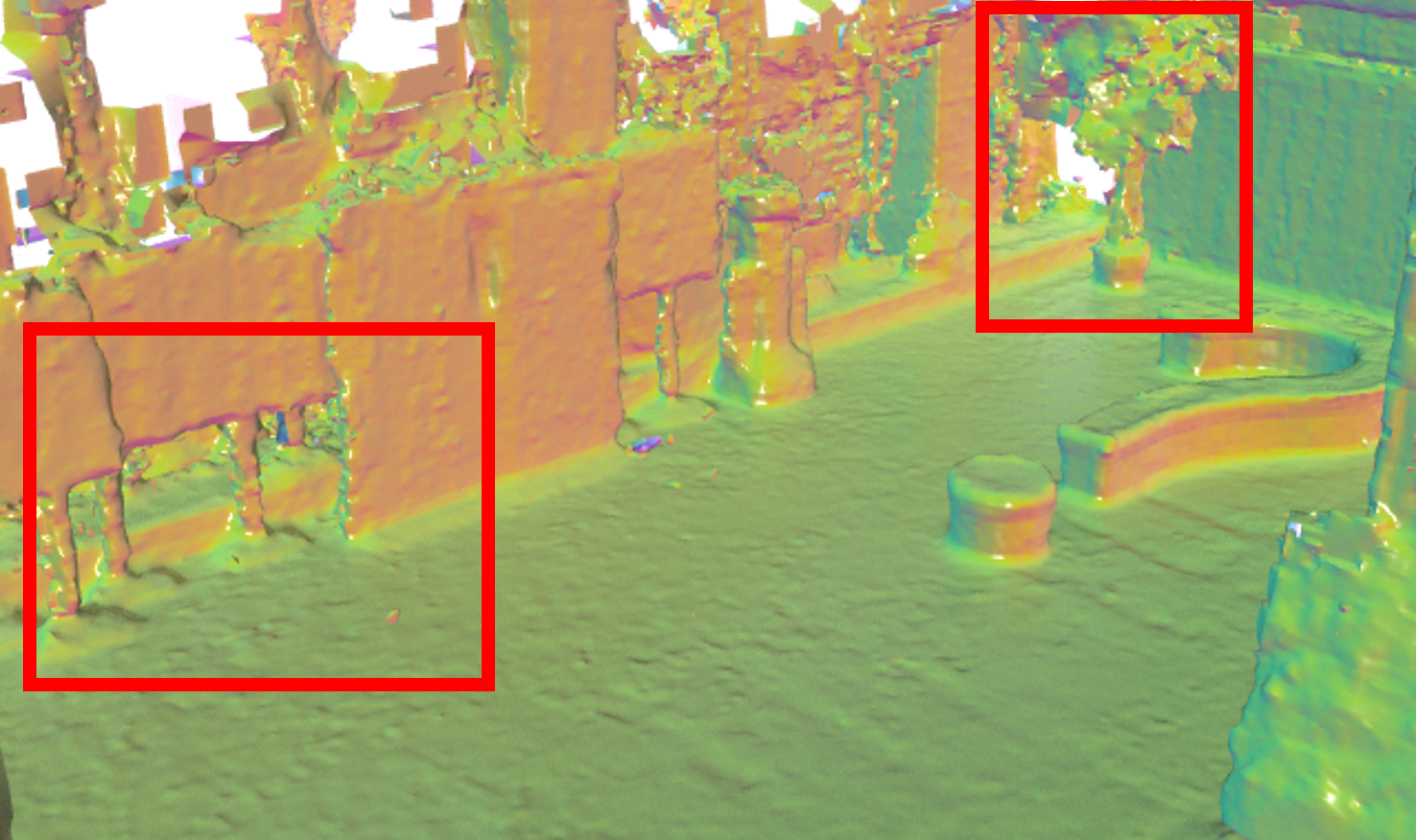}\vspace{4pt}
\includegraphics[height=1.8cm]{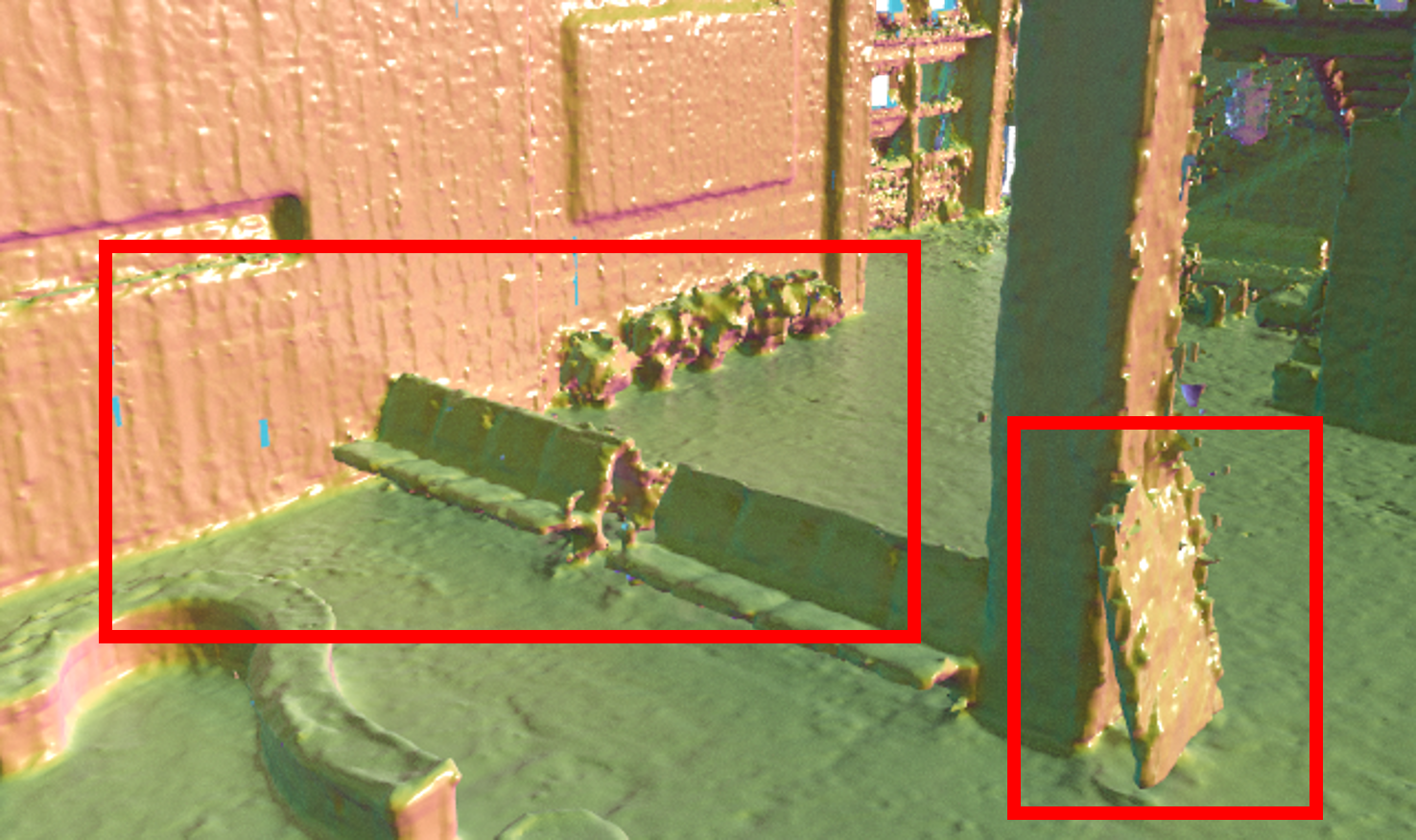}
\end{minipage}}
\subfigure[Voxblox]{
\begin{minipage}[b]{0.34\linewidth}
\includegraphics[height=1.8cm]{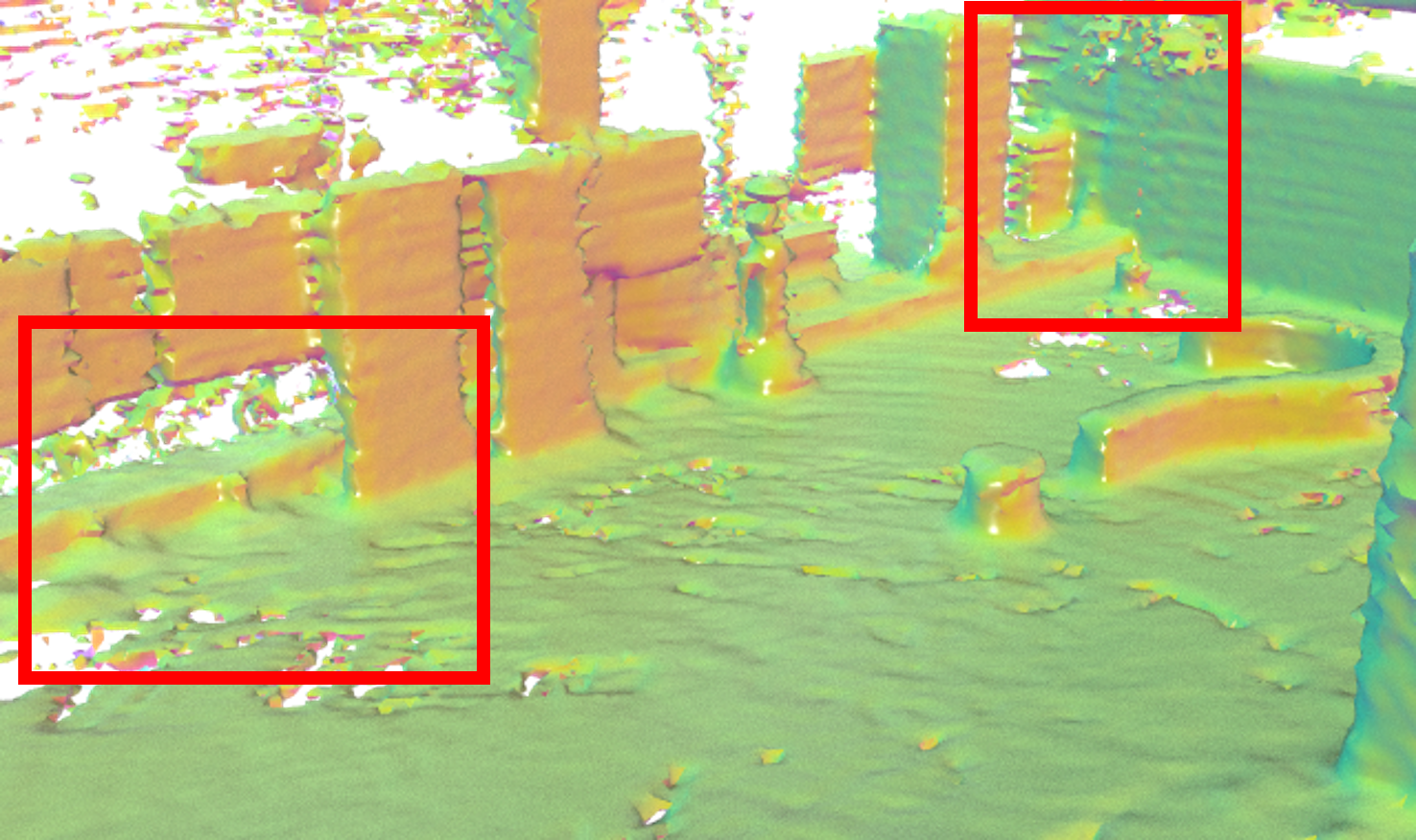}\vspace{4pt}
\includegraphics[height=1.8cm]{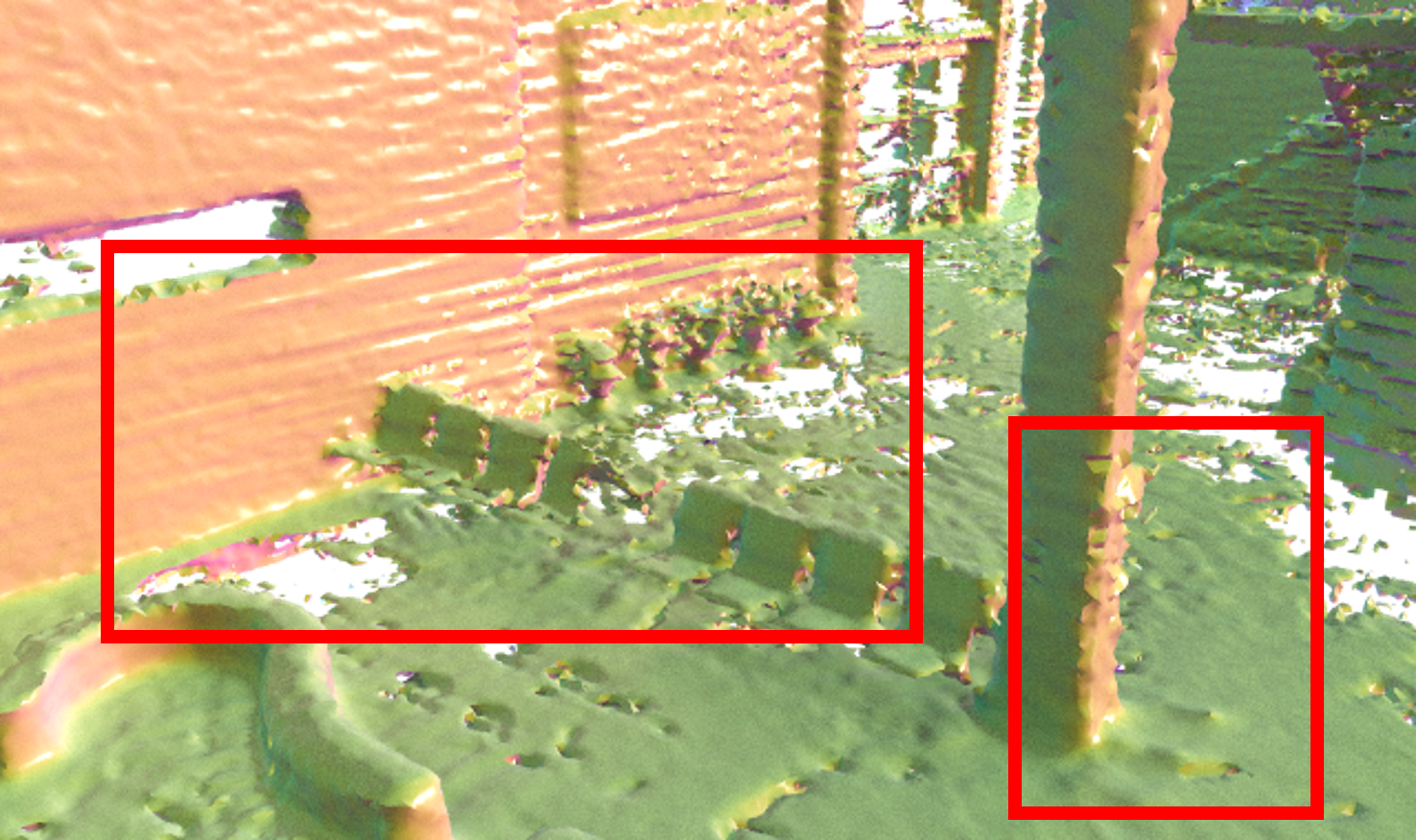}
\end{minipage}}
\caption{
Case comparison of different methods on the lobby dataset. Each row shows the results of a case area using different methods. Bulletin board pole, plant, magazine rack, and floor are highlighted in the red box.
}
\label{fig:lobby}
\end{figure}

\begin{figure}[H]
\renewcommand{\thefigure}{2}
\centering
\subfigure[Dataset]{
\begin{minipage}[b]{0.26\linewidth}
\includegraphics[height=1.8cm]{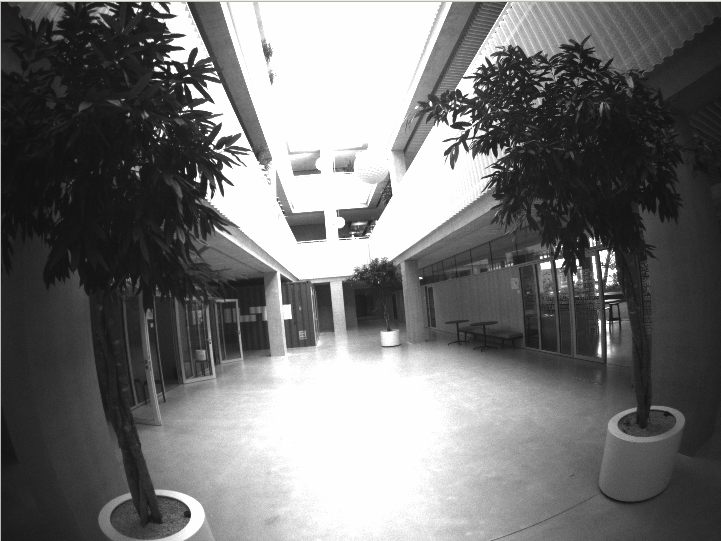}\vspace{4pt}
\includegraphics[height=1.8cm]{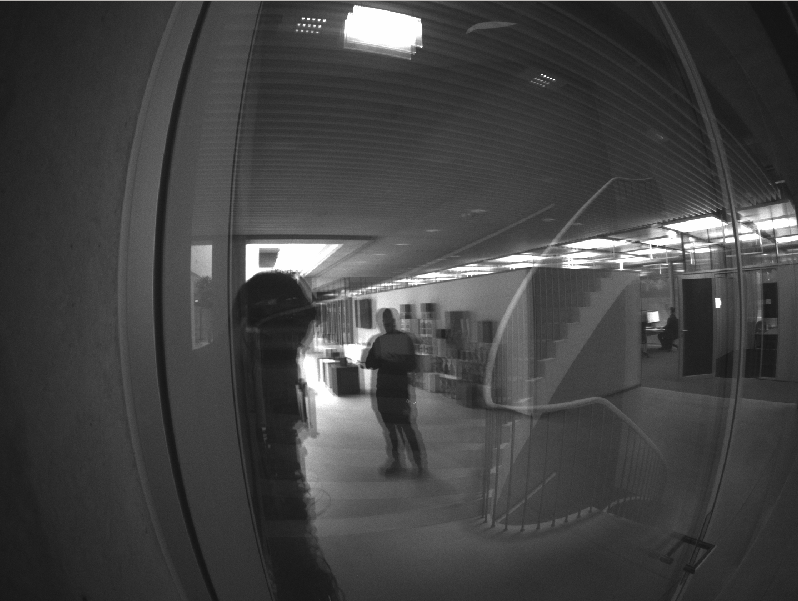}
\end{minipage}}
\subfigure[Ours]{
\begin{minipage}[b]{0.33\linewidth}
\includegraphics[height=1.8cm]{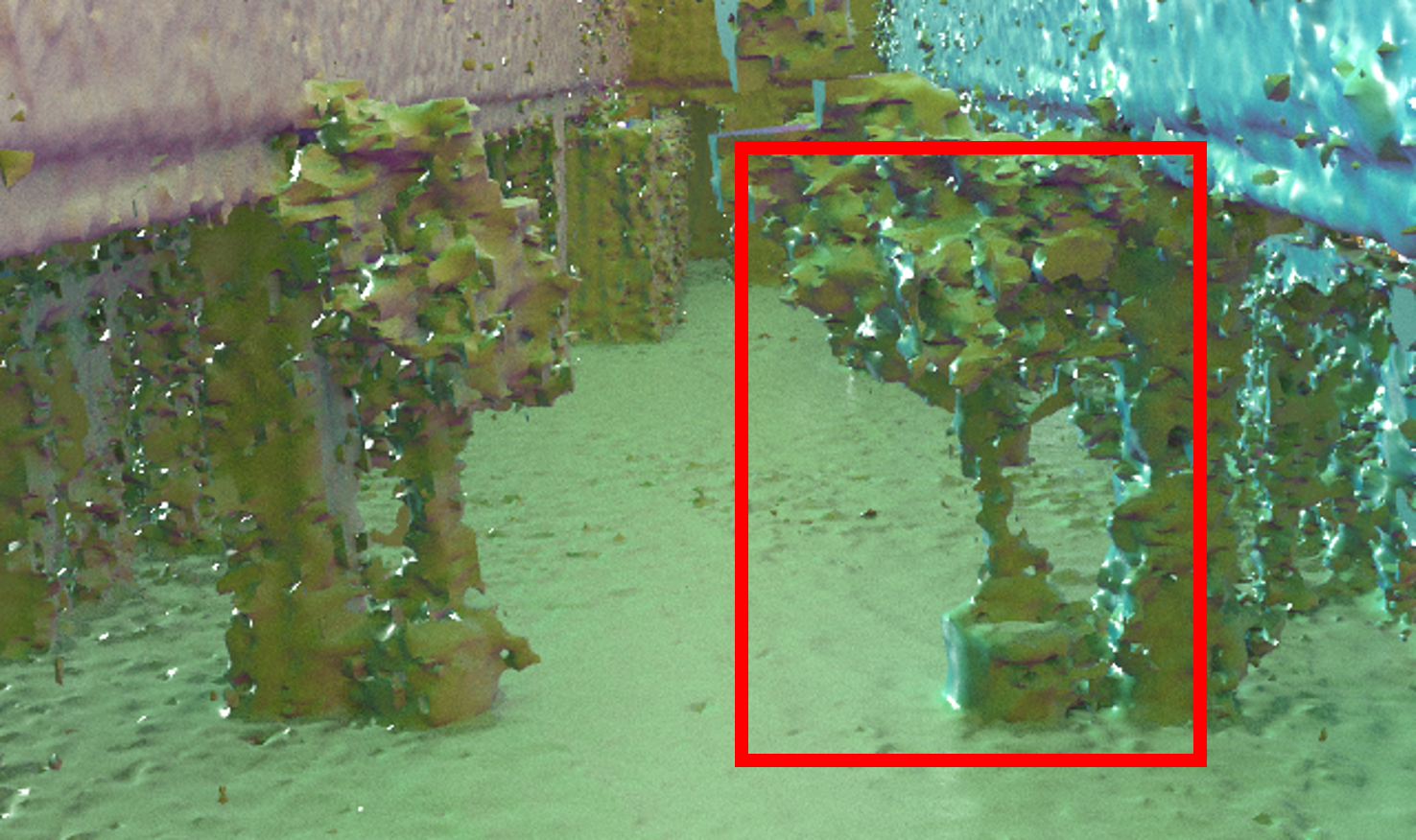}\vspace{4pt}
\includegraphics[height=1.8cm]{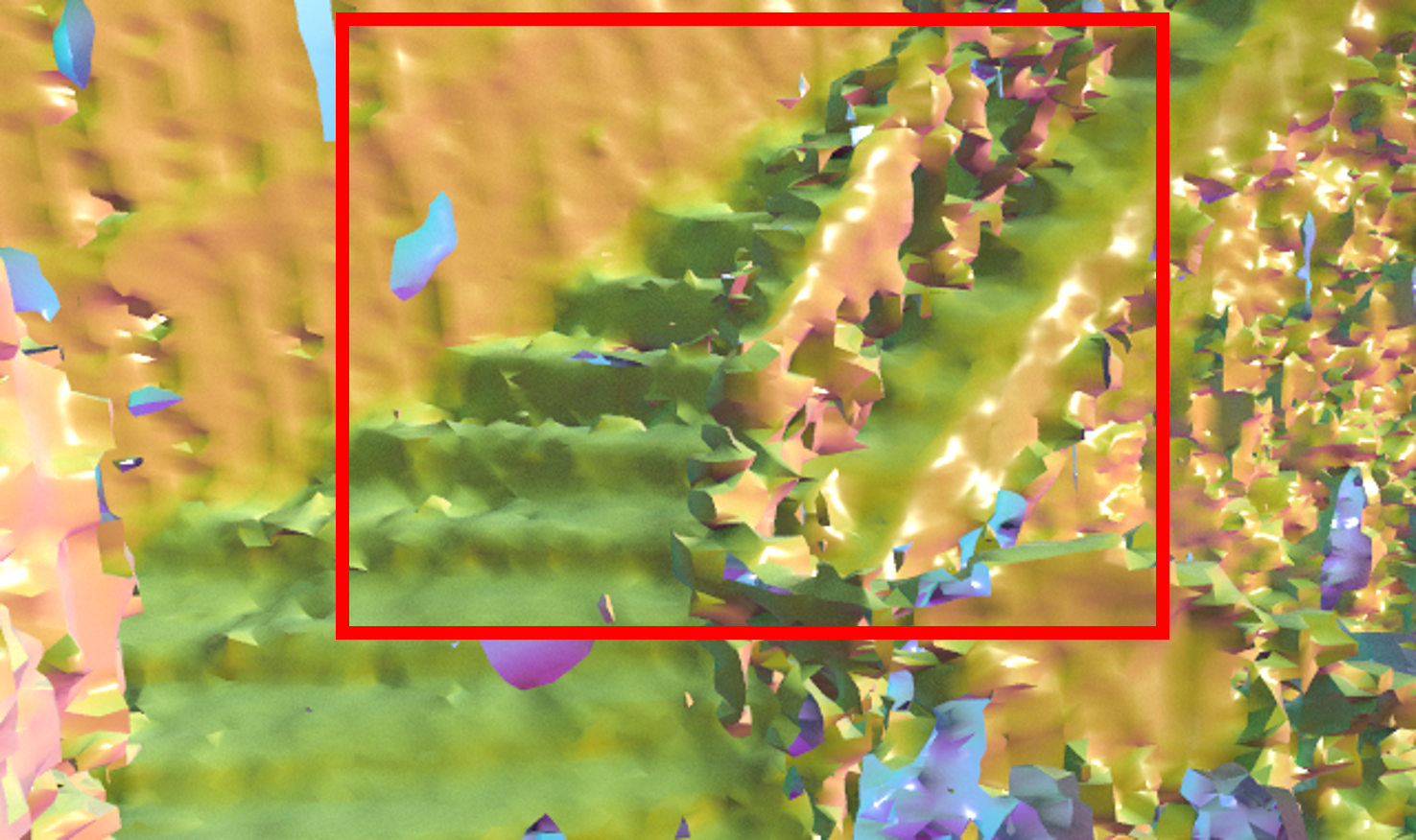}
\end{minipage}}
\subfigure[Voxblox]{
\begin{minipage}[b]{0.34\linewidth}
\includegraphics[height=1.8cm]{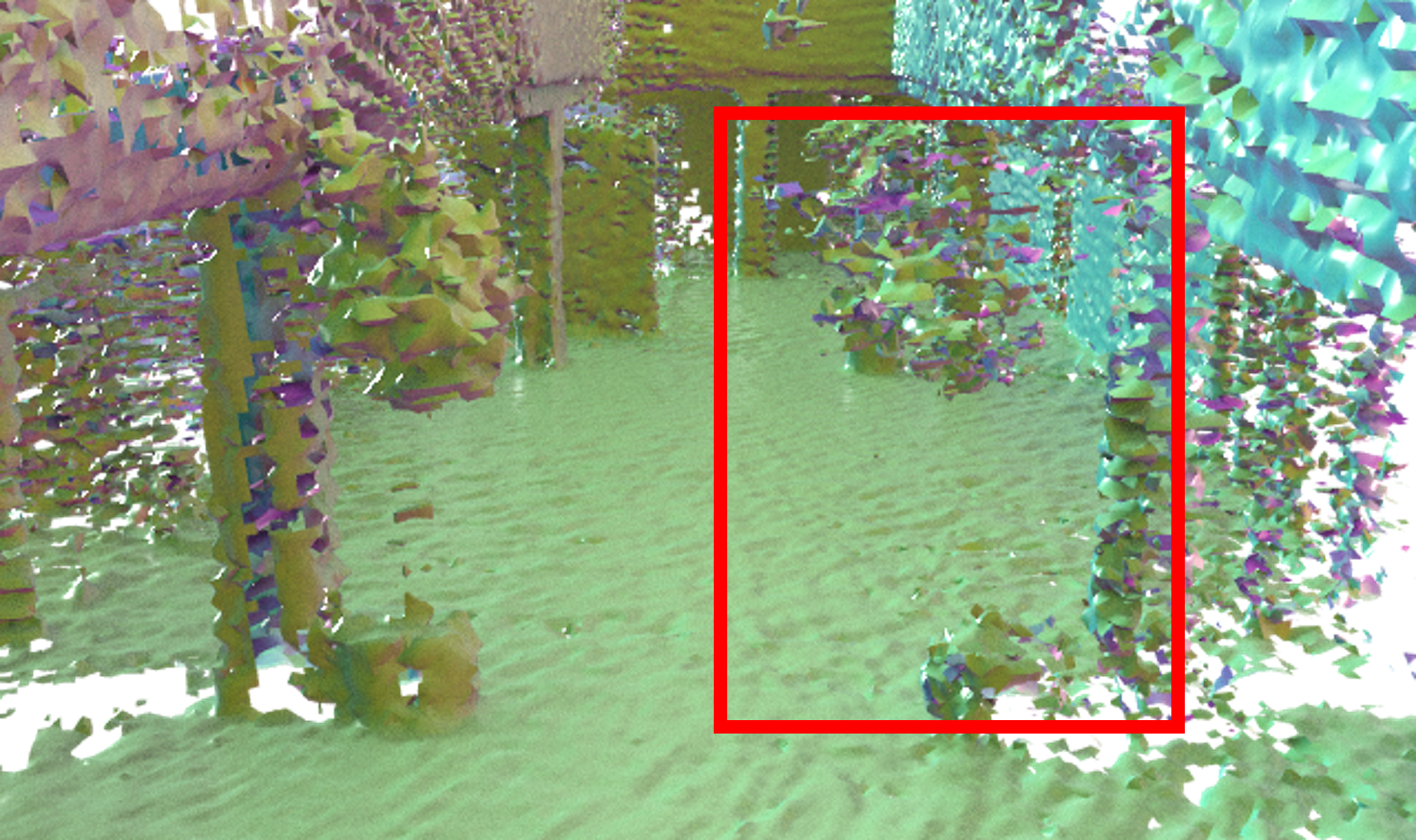}\vspace{4pt}
\includegraphics[height=1.8cm]{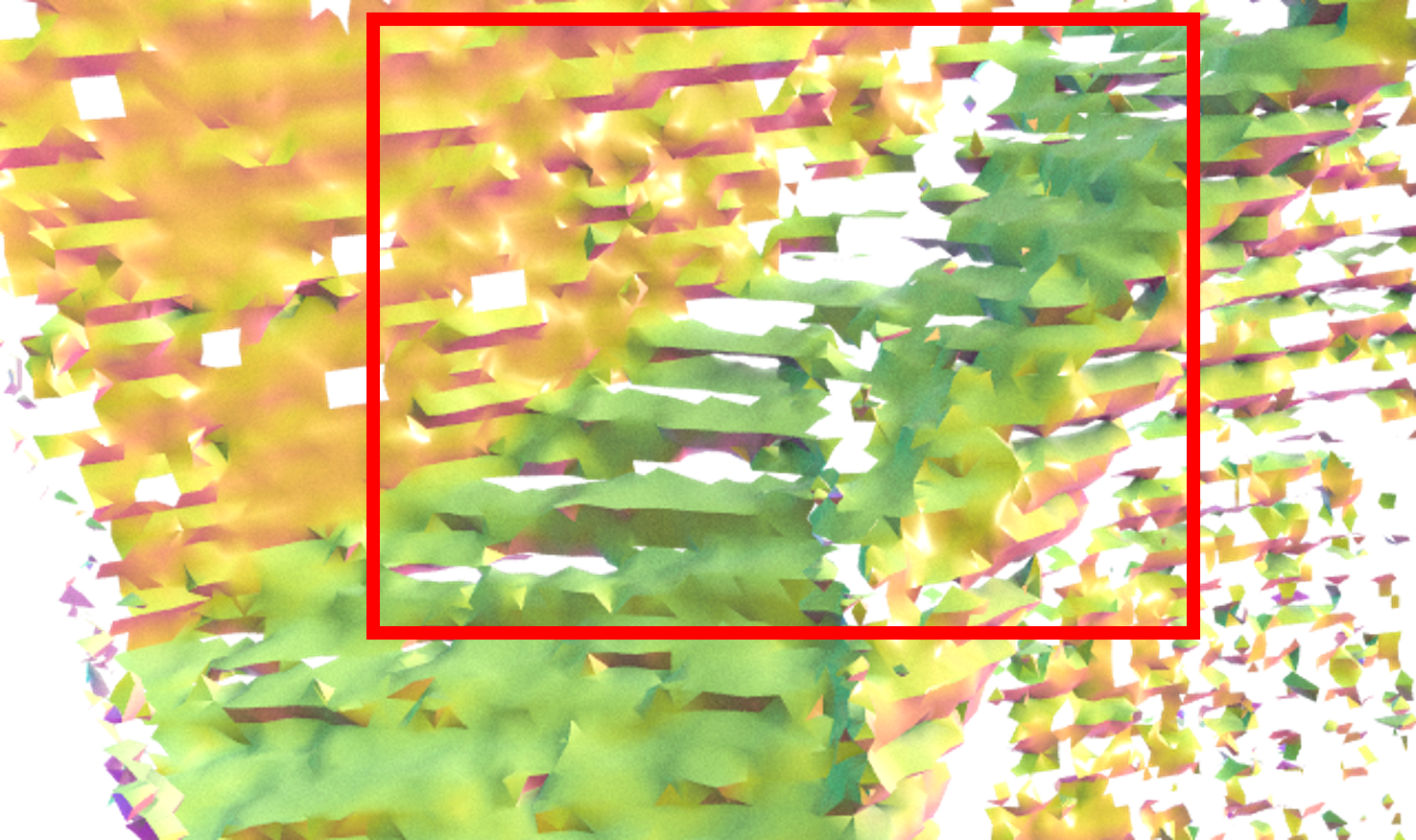}
\end{minipage}}
\caption{Case comparison of different methods on the Hilti SLAM 2021 Office. Each row shows the results of a case area using different methods. Plant and stairs are highlighted in the red box.}
\label{fig:hilti}
\end{figure}

\textbf{Quantitative Analysis:}
The reconstruction from the sequential point cloud of the lobby dataset and Hilti SLAM 2021 are shown in Fig.~\ref{fig:lobby} and Fig.~\ref{fig:hilti} qualitatively. Quantitative evaluations with comparison methods are reported in Tab.~\ref{tab:indoor}, which shows that our method has a better performance on chamfer distance than the Voxblox \cite{voxblox}. It can be seen that our method can reconstruct the details of indoor objects, such as the poles of bulletin boards, tree potted plants, etc., and can also reconstruct the smooth and complete surface of objects.
\begin{table}[H]
\renewcommand{\thetable}{I}
    \centering
    \setlength{\tabcolsep}{4pt}
    \renewcommand\arraystretch{1.2}
    \caption{\label{tab:indoor}MAPPING QUALITY ON INDOOR DATASETS}
    \begin{tabular}{cccccccl}
        \toprule
             \textbf{Dataset} & \textbf{Method} & \textbf{Comp.}$\downarrow$ & \textbf{Acc.}$\downarrow$ & \textbf{C-L1}$\downarrow$  & \textbf{Re.}$\uparrow$ & \textbf{Pre.}$\uparrow$ & \textbf{F1}$\uparrow$\\
        \midrule
            \multirow{2}*{Lobby} & Voxblox & 4.69 & \textbf{1.14} & 2.91 & 78.46 & \textbf{99.89} & 87.89\\
                & \textbf{Ours} & \textbf{3.25} & 1.36 & \textbf{2.31} & \textbf{89.42} & 99.59 & \textbf{94.24}\\
        \midrule
            \multirow{2}*{Hilti} & Voxblox & 14.94  & \textbf{2.77}  &  8.86 & 29.17 & \textbf{86.53}  & 43.64\\
                & \textbf{Ours} & \textbf{4.72} & 2.92 & \textbf{3.82} & \textbf{65.97} & 83.59 & \textbf{73.74}\\ 
        \bottomrule
    \end{tabular}
\end{table}


\end{document}